\documentclass[journal,onecolumn]{IEEEtran}
\usepackage{url}
\usepackage[numbers]{natbib}
\usepackage{graphicx}
\usepackage{amsmath, bm}
\usepackage{url}
\usepackage{soul}
\usepackage{multirow}
\usepackage{caption}
\usepackage{subcaption}
\usepackage{wrapfig}
\usepackage{amssymb}
\usepackage{comment}
\usepackage{algorithmic,algorithm}

\algsetup{indent=1.5em} 
\usepackage{natbib}

\usepackage[
  separate-uncertainty = true,
  multi-part-units = repeat
]{siunitx}

\begin{document}

\title{Recent Trends in Deep Learning Based\\Natural Language Processing}

\author{Tom Young$^{\dagger\equiv}$, Devamanyu Hazarika$^{\ddagger\equiv}$, Soujanya Poria$^{\oplus\equiv}$\thanks{$^{\equiv}$ means authors contributed equally}, Erik Cambria$^{\bigtriangledown*}$\thanks{$^{*}$ Corresponding author (e-mail: cambria@ntu.edu.sg)}\\
	
	\vspace{5mm}
	\small{$^{\dagger}$ School of Information and Electronics, Beijing Institute of Technology, China}\\
	\small{$^{\ddagger}$ School of Computing, National University of Singapore, Singapore}\\
	\small{$^{\oplus}$ Temasek Laboratories, Nanyang Technological University, Singapore}\\
	\small{$^{\bigtriangledown}$ School of Computer Science and Engineering, Nanyang Technological University, Singapore}\\
	
}

\maketitle

\begin{abstract}

Deep learning methods employ multiple processing layers to learn hierarchical representations of data, and have produced state-of-the-art results in many domains. Recently, a variety of model designs and methods have blossomed in the context of natural language processing (NLP). In this paper, we review significant deep learning related models and methods that have been employed for numerous NLP tasks and provide a walk-through of their evolution. We also summarize, compare and contrast the various models and put forward a detailed understanding of the past, present and future of deep learning in NLP.

\end{abstract}


\IEEEpeerreviewmaketitle

\section{Introduction}
Natural language processing
(NLP) is a theory-motivated
range of computational techniques
for the automatic analysis and
representation of human language.
NLP research has evolved from the era
of punch cards and batch processing, in
which the analysis of a sentence could
take up to 7 minutes, to the era of
Google and the likes of it, in which
millions of webpages can be processed
in less than a second~\cite{camjum}. NLP enables computers to perform a wide range of natural language related tasks at all levels, ranging from parsing and part-of-speech (POS) tagging, to machine translation and dialogue systems. 

\begin{figure}[h]
	\includegraphics[width=0.5\linewidth]{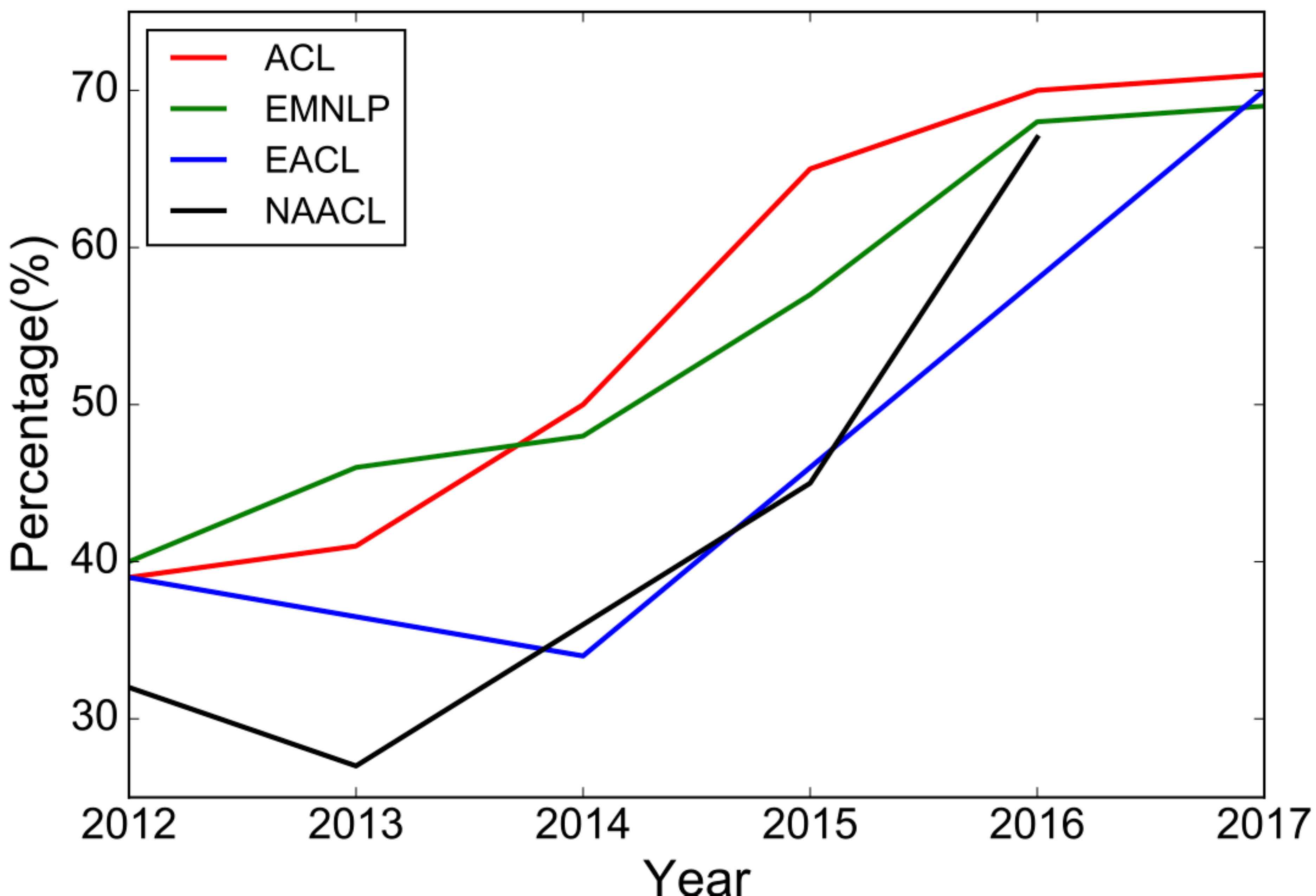}
	\centering
	\caption{Percentage of deep learning papers in ACL, EMNLP, EACL, NAACL over the last 6 years (long papers).}
	\label{fig:distributional}
\end{figure}

Deep learning architectures and algorithms have already made
impressive advances in fields such as computer vision and pattern
recognition. Following this trend, recent NLP research is now increasingly
focusing on the use of new deep learning methods (see Figure~\ref{fig:distributional}). For decades, machine learning approaches targeting NLP problems have been based on shallow models (e.g., SVM and logistic regression) trained on very high dimensional and sparse features. In the last few years, neural networks based on dense vector representations have been producing superior results on various NLP tasks. This trend is sparked by the success of word embeddings~\cite{mikolov2010recurrent,mikolov2013distributed} and deep learning methods~\cite{socher2013recursive}. Deep learning enables multi-level automatic feature representation learning. In contrast, traditional machine learning based NLP systems liaise heavily on hand-crafted features. Such hand-crafted features are time-consuming and often incomplete. 

\citet{collobert2011natural} demonstrated that a simple deep learning framework outperforms
most state-of-the-art approaches in several NLP tasks such
as named-entity recognition (NER), semantic role labeling (SRL),
and POS tagging. Since then, numerous complex deep learning based algorithms have been proposed to solve difficult NLP tasks.
We review major deep learning related models and methods applied to natural language tasks such as convolutional neural networks (CNNs), recurrent neural networks (RNNs), and recursive neural networks. We also discuss memory-augmenting strategies, attention mechanisms and how unsupervised models, reinforcement learning methods and recently, deep generative models have been employed for language-related tasks. 

To the best of our knowledge, this work is the first of its type to comprehensively cover the most popular deep learning methods in NLP research today~\footnote{We intend to update this article with time as and when significant advances are proposed and used by the community}. 
The work by~\citet{goldberg2016primer} only presented the basic principles for applying neural networks to NLP in a tutorial manner. We believe this paper will give readers a more comprehensive idea of current practices in this domain.


The structure of the paper is as follows: Section~\ref{sec:2} introduces the concept of distributed representation, the basis of sophisticated deep learning models; next, Sections~\ref{sec:3}, \ref{sec:4}, and \ref{sec:5} discuss popular models such as convolutional, recurrent, and recursive neural networks, as well as their use in various NLP tasks; following, Section~\ref{sec:6} lists recent applications of reinforcement learning in NLP and new developments in unsupervised sentence representation learning; later, Section~\ref{sec:7} illustrates the recent trend of coupling deep learning models with memory modules; finally, Section~\ref{sec:8} summarizes the performance of a series of deep learning methods on standard datasets about major NLP topics.

\section{Distributed Representation}\label{sec:2}
Statistical NLP has emerged as the primary option for modeling complex natural language tasks. However, in its beginning, it often used to suffer from the notorious \textit{curse of dimensionality} while learning joint probability functions of language models. This led to the motivation of learning distributed representations of words existing in low-dimensional space~\cite{bengio2003neural}. 

\subsection{Word Embeddings}
\begin{figure}[b]
	\includegraphics[width=0.5\linewidth]{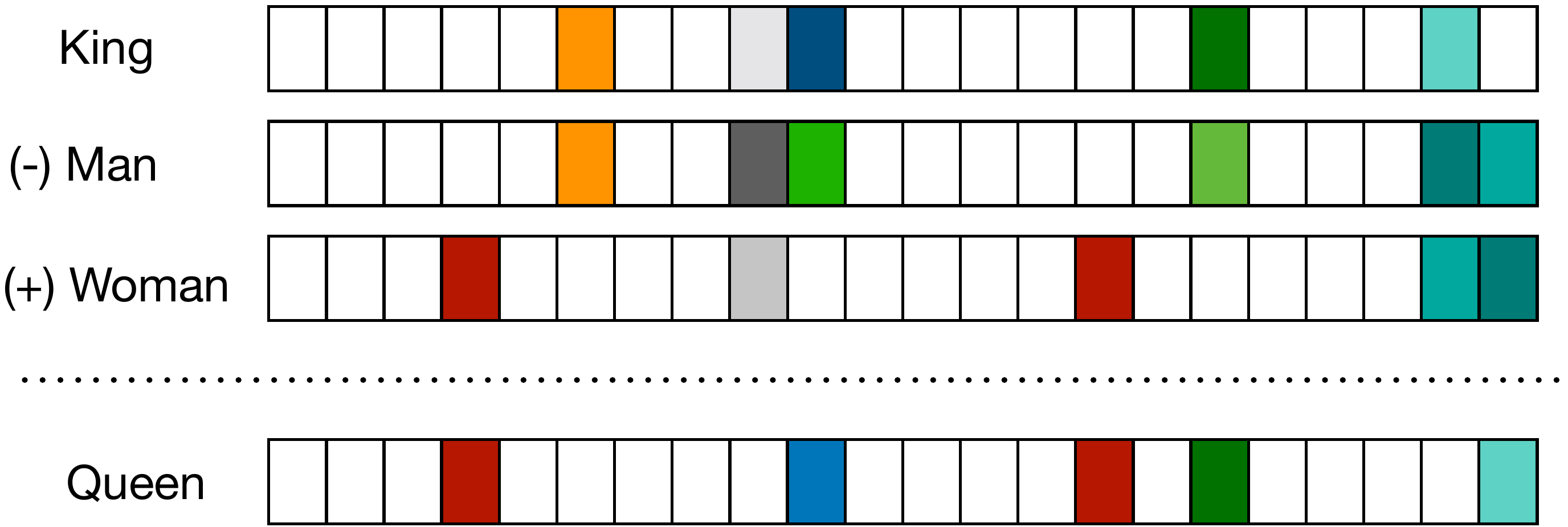}
	\centering
	\caption{Distributional vectors represented by a ${\bf D}$-dimensional vector where ${\bf D} << {\bf V} $, where ${\bf V} $ is size of Vocabulary. Figure Source: \protect\url{http://veredshwartz.blogspot.sg.}}
	\label{fig:distributional2}
\end{figure}

Distributional vectors or word embeddings (Fig.~\ref{fig:distributional2}) essentially follow the distributional hypothesis, according to which words with similar meanings tend to occur in similar context. Thus, these vectors try to capture the characteristics of the neighbors of a word.
The main advantage of distributional vectors is that they capture similarity between words. Measuring similarity between vectors is possible, using measures such as cosine similarity.
Word embeddings are often used as the first data processing layer in a deep learning model. Typically, word embeddings are pre-trained by optimizing an auxiliary objective in a large unlabeled corpus, such as predicting a word based on its context~\cite{mikolov2013efficient,mikolov2013distributed}, where the learned word vectors can capture general syntactical and semantic information. Thus, these embeddings have proven to be efficient in capturing context similarity, analogies and due to its smaller dimensionality, are fast and efficient in processing core NLP tasks.

Over the years, the models that create such embeddings have been shallow neural networks and there has not been need for deep networks to create good embeddings. However, deep learning based NLP models invariably represent their words, phrases and even sentences using these embeddings. This is in fact a major difference between traditional word count based models and deep learning based models.
Word embeddings have been responsible for state-of-the-art results in a wide range of NLP tasks~\cite{weston2011wsabie,socher2011parsing,turney2010frequency,camsui}. 

\begin{figure}[ht]
	\includegraphics[width=0.5\linewidth]{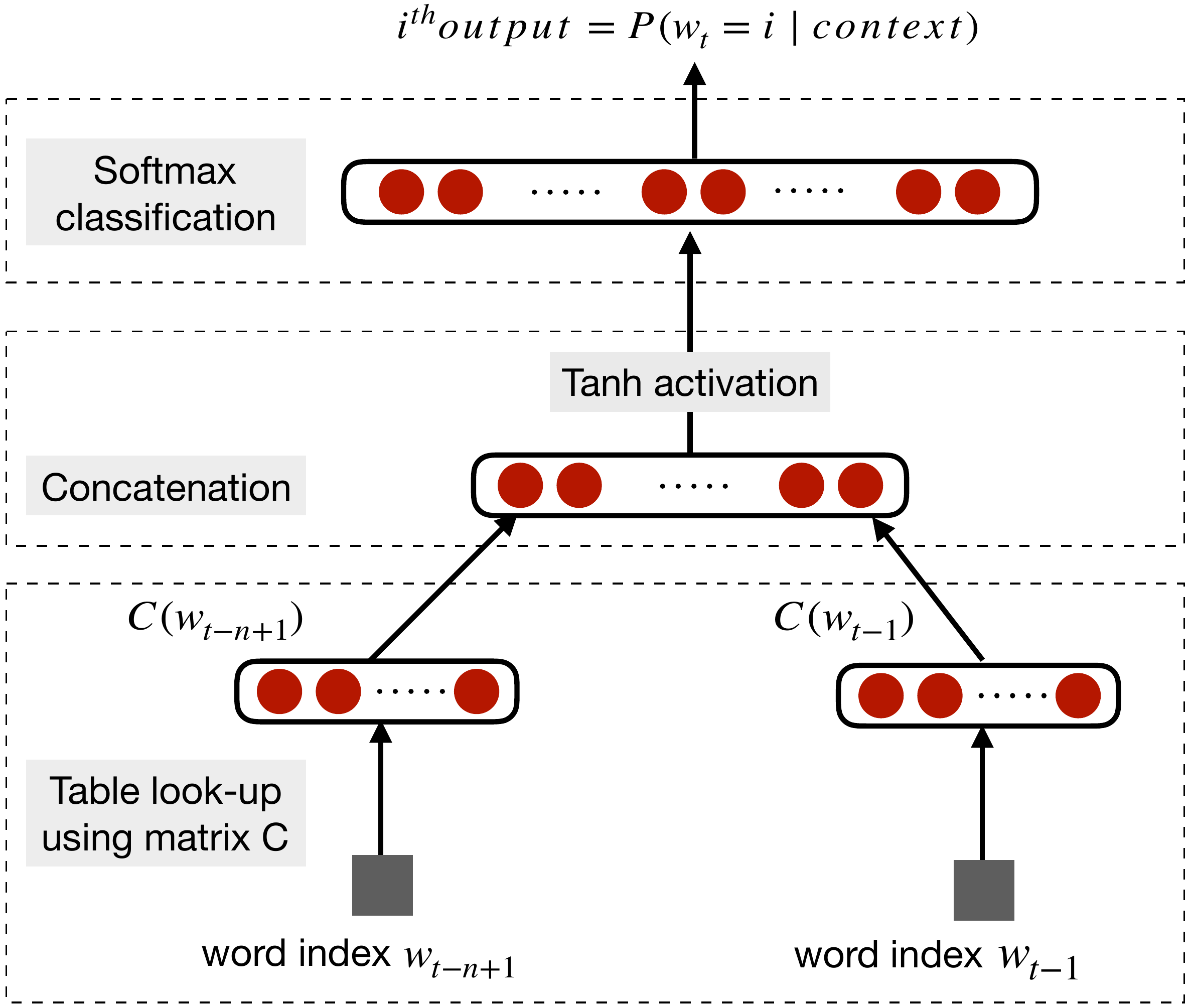}
	\centering
	\caption{Neural Language Model (Figure reproduced from~\citet{bengio2003neural}). $C(i)$ is the $i^{th}$ word embedding.}
	\label{fig:NeuralModel}
\end{figure}

For example, \citet{glorot2011domain} used embeddings along with stacked denoising autoencoders for domain adaptation in sentiment
classification and \citet{hermann2013role} presented combinatory categorial autoencoders to learn the compositionality of sentence. Their wide usage across the recent literature shows their effectiveness and importance in any deep learning model performing a NLP task.

Distributed representations (embeddings) are mainly learned through context. During 1990s, several research developments~\cite{elman1991distributed} marked the foundations of research in distributional semantics. A more detailed summary of these early trends is provided in~\cite{glenberg2000symbol,dumais2004latent}. Later developments were adaptations of these early works, which led to creation of topic models like latent Dirichlet allocation~\cite{blei2003latent} and language models~\cite{bengio2003neural}. These works laid out the foundations of representation learning in natural language.

In 2003, \citet{bengio2003neural} proposed a neural language model which learned distributed representations for words (Fig.~\ref{fig:NeuralModel}). Authors argued that these word representations, once compiled into sentence representations using joint probability of word sequences, achieved an exponential number of semantically neighboring sentences. This, in turn, helped in generalization since unseen sentences could now gather higher confidence if word sequences with similar words (in respect to nearby word representation) were already seen. 

\citet{collobert2008unified} were the first work to show the utility of pre-trained word embeddings. They proposed a neural network architecture that forms the foundation to many current approaches. The work also establishes word embeddings as a useful tool for NLP tasks. However, the immense popularization of word embeddings was arguably due to~\citet{mikolov2013distributed} who proposed the continuous bag-of-words (CBOW) and skip-gram models to efficiently construct high-quality distributed vector representations. Propelling their popularity was the unexpected side effect of the vectors exhibiting compositionality, i.e., adding two word vectors results in a vector that is a semantic composite of the individual words, e.g., `man' + `royal' = `king'. The theoretical justification for this behavior was recently given by~\citet{gittens2017skip}, which stated that compositionality is seen only when certain assumptions are held, e.g., the assumption that words need to be uniformly distributed in the embedding space. 

Glove by \citet{pennington2014glove} is another famous word embedding method which is essentially a ``count-based" model. Here, the word co-occurrence count matrix is pre-processed by normalizing the counts and log-smoothing operation. This matrix is then factorized to get lower dimensional representations which is done by minimizing a ``reconstruction loss".

Below, we provide a brief description of the word2vec method proposed by~\citet{mikolov2013distributed}. 

\subsection{Word2vec}
Word embeddings were revolutionized by~\citet{mikolov2013efficient,mikolov2013distributed} who proposed the CBOW and skip-gram models. CBOW computes the conditional probability of a target word given the context words surrounding it across a window of size $k$. 
On the other hand, the skip-gram model does the exact opposite of the CBOW model, by predicting the surrounding context words given the central target word. The context words are assumed to be located symmetrically to the target words within a distance equal to the window size in both directions.
In unsupervised settings, the word embedding dimension is determined by the accuracy of prediction. As the embedding dimension increases, the accuracy of prediction also increases until it converges at some point, which is considered the optimal embedding dimension as it is the shortest without compromising accuracy.

Let us consider a simplified version of the CBOW model where only one word is considered in the context. This essentially replicates a bigram language model. 

\begin{figure}[h]
	\includegraphics[width=0.5\linewidth]{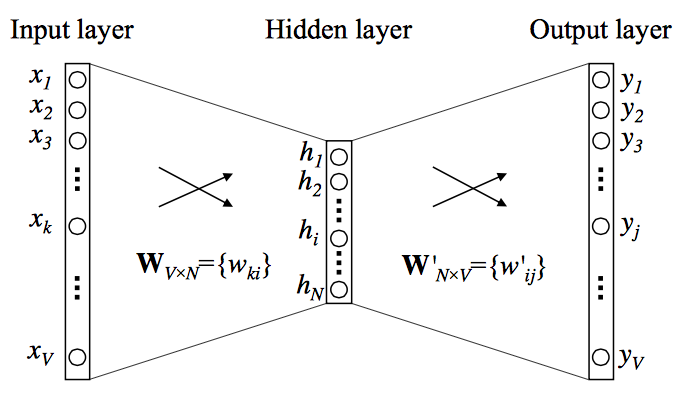}
	\centering
	\caption{Model for CBOW (Figure source:~\citet{rong2014word2vec})}
	\label{fig:CBOW} 
\end{figure}

As shown in Fig.~\ref{fig:CBOW}, the CBOW model is a simple fully connected neural network with one hidden layer. The input layer, which takes the one-hot vector of context word has ${V}$ neurons while the hidden layer has ${N}$ neurons. The output layer is softmax probability over all words in the vocabulary. The layers are connected by weight matrix 
${\bf W} \in \mathcal{R}^{V \times N } $ and ${\bf W^{'}} \in \mathcal{R}^{H \times V}$, respectively. Each word from the vocabulary is finally represented as two learned vectors $ {\bf v_c} $ and ${\bf v_w}$, corresponding to context and target word representations, respectively. Thus, $k^{th}$ word in the vocabulary will have
\begin{gather}
{\bf v_c} = {\bf W_{(k,.)}} \quad and \quad {\bf v_w} = {\bf W^{'}_{(.,k)}}
\end{gather}

Overall, for any word $w_i$ with given context word $c$ as input, 
\begin{gather}
 {p\left(\frac{w_i}{c}\right)} = {\bf y_i} = \frac{e^{u_i}}{\sum_{i=1}^{V} e^{u_i}} \quad where,\ \ {u_i} = {\bf v_{w_i}^T.v_c}
\end{gather}
The parameters $ {\bf \theta} = {\bf \{v_w, v_c\}}_{w,c \,\in\, \text{Vocab}}$ are learned by defining the objective function as the log-likelihood and finding its gradient as
\begin{gather}
{\bf l(\theta)} = {\bf \sum_{w \in Vocab} log\left(p\left(\frac{w}{c}\right)\right)} \\
{\bf \frac{\partial l(\theta)}{\partial v_w}} = {\bf v_c\left(1-p\left(\frac{w}{c}\right)\right)}
\end{gather}

In the general CBOW model, all the one-hot vectors of context words are taken as input simultaneously, i.e,
\begin{equation}
{\bf h} = {\bf W^T(x_1 + x_2 + ... + x_c)}
\end{equation}

One limitation of individual word embeddings is their inability to represent phrases~\cite{mikolov2013distributed}, where the combination of two or more words -- e.g., idioms like ``hot potato'' or named entities such as ``Boston Globe' -- does not represent the combination of meanings of individual words. One solution to this problem, as explored by~\citet{mikolov2013distributed}, is to identify such phrases based on word co-occurrence and train embeddings for them separately. Later methods have explored directly learning n-gram embeddings from unlabeled data~\cite{johnson2015semi}. 

Another limitation comes from learning embeddings based only on a small window of surrounding words, sometimes words such as \emph{good} and \emph{bad} share almost the same embedding~\cite{socher2011semi}, which is problematic if used in tasks such as sentiment analysis~\cite{wang2015predicting}. At times these embeddings cluster semantically-similar words which have opposing sentiment polarities. This leads the downstream model used for the sentiment analysis task to be unable to identify this contrasting polarities leading to poor performance. \citet{tang2014learning} addressed this problem by proposing sentiment specific word embedding (SSWE). Authors incorporated the supervised sentiment polarity of text in their loss functions while learning the embeddings.

A general caveat for word embeddings is that they are highly dependent on the applications in which it is used. \citet{labutov2013re} proposed task specific embeddings which retrain the word embeddings to align them in the current task space. This is very important as training embeddings from scratch requires large amount of time and resource. \citet{mikolov2013efficient} tried to address this issue by proposing \textit{negative sampling} which does frequency-based sampling of negative terms while training the word2vec model. 

Traditional word embedding algorithms assign a distinct vector to each word. This makes them unable to account for polysemy. In a recent work, \citet{upadhyay2017beyond} provided an innovative way to address this deficit. The authors leveraged multilingual parallel data to learn multi-sense word embeddings. 
For example, the English word bank, when translated to French provides two different words: \textit{banc} and \textit{banque} representing financial and geographical meanings, respectively. Such multilingual distributional information helped them in accounting for polysemy.

Table~\ref{tab:embedding} provides a directory of existing frameworks that are frequently used for creating embeddings which are further incorporated into deep learning models.

\begin{table}[h]
	\small
	\begin{center}
		\begin{tabular}{c|c|c}
			\hline
			Framework & Language & URL \\ \hline
			S-Space & Java & \scriptsize{\url{https://github.com/fozziethebeat/S-Space}}\\
			Semanticvectors & Java &\scriptsize{\url{https://github.com/semanticvectors/}}\\
			Gensim & Python &\scriptsize{\url{https://radimrehurek.com/gensim/}}\\
			Pydsm & Python &\scriptsize{\url{https://github.com/jimmycallin/pydsm}}\\
			Dissect & Python &\scriptsize{\url{http://clic.cimec.unitn.it/composes/toolkit/}}\\
      FastText & Python &\scriptsize{\url{https://fasttext.cc/}}\\
      Elmo & Python & \scriptsize{\url{https://tfhub.dev/google/elmo/2}}\\
			\hline
		\end{tabular}
	\end{center}
	
	\caption {Frameworks providing word embedding tools and methods.}
	\label{tab:embedding}
\end{table}

\subsection{Character Embeddings}
Word embeddings are able to capture syntactic and semantic information, yet for tasks such as POS-tagging and NER, intra-word morphological and shape information can also be very useful. Generally speaking, building natural language understanding systems at the character level has attracted certain research attention~\cite{kim2016character,dos2014deep,santos2015boosting,santos2014learning}. Better results on morphologically rich languages are reported in certain NLP tasks. \citet{santos2015boosting} applied character-level representations, along with word embeddings for NER, achieving state-of-the-art results in Portuguese and Spanish corpora. \citet{kim2016character} showed positive results on building a neural language model using only character embeddings. \citet{maalab} exploited several embeddings, including character trigrams, to incorporate prototypical and hierarchical information for learning pre-trained label embeddings in the context of NER.

A common phenomenon for languages with large vocabularies is the unknown word issue, also known as \textit{out-of-vocabulary} (OOV) words. Character embeddings naturally deal with it since each word is considered as no more than a composition of individual letters. In languages where text is not composed of separated words but individual characters and the semantic meaning of words map to its compositional characters (such as Chinese), building systems at the character level is a natural choice to avoid word segmentation~\cite{chen2015joint}. Thus, works employing deep learning applications on such languages tend to prefer character embeddings over word vectors~\cite{zheng2013deep}. For example, \citet{penrad} proved that radical-level processing could greatly improve sentiment classification performance. In particular, the authors proposed two types of Chinese radical-based hierarchical embeddings, which incorporate not only semantics at radical and character level, but also sentiment information. \citet{bojanowski2016enriching} also tried to improve the representation of words by using character-level information in morphologically-rich languages. They approached the skip-gram method by representing words as bag-of-character n-grams. Their work thus had the effectiveness of the skip-gram model along with addressing some persistent issues of word embeddings. The method was also fast, which allowed training models on large corpora quickly. Popularly known as \textit{FastText}, such a method stands out over previous methods in terms of speed, scalability, and effectiveness. 

Apart from character embeddings, different approaches have been proposed for OOV handling. \citet{herbelot2017high} provided on-the-fly OOV handling by initializing the unknown words as the sum of the context words and refining these words with a high learning rate. However, their approach is yet to be tested on typical NLP tasks. \citet{pinter2017mimicking} provided an interesting approach of training a character-based model to recreate pre-trained embeddings. This allowed them to learn a compositional mapping form character to word embedding, thus tackling the OOV problem. 

Despite the ever growing popularity of distributional vectors, recent discussions on their relevance in the long run have cropped up. For example, \citet{lucy2017distributional} has recently tried to evaluate how well the word vectors capture the necessary facets of conceptual meaning. The authors have discovered severe limitations in perceptual understanding of the concepts behind the words, which cannot be inferred from distributional semantics alone. A possible direction for mitigating these deficiencies will be grounded learning, which has been gaining popularity in this research domain.




\subsection{Contextualized Word Embeddings} \label{sec:contextualembeddings}

The quality of word representations is generally gauged by its ability to encode syntactical information and handle polysemic behavior (or word senses). These properties result in improved semantic word representations. Recent approaches in this area encode such information into its embeddings by leveraging the context. These methods provide deeper networks that calculate word representations as a function of its context. 

Traditional word embedding methods such as Word2Vec and Glove consider all the sentences where a word is present in order to create a global vector representation of that word. However, a word can have completely different senses or meanings in the contexts. For example, lets consider these two sentences - 1) ``The \textit{bank} will not be accepting cash on Saturdays'' 2) ``The river overflowed the \textit{bank}.''. The word senses of \textit{bank} are different in these two sentences depending on its context.
Reasonably, one might want two different vector representations of the word \textit{bank} based on its two different word senses. The new class of models adopt this reasoning by diverging from the concept of global word representations and proposing contextual word embeddings instead.

Embedding from Language Model (ELMo) \cite{peters2018deep} is one such method that provides deep contextual embeddings. ELMo produces word embeddings for each context where the word is used, thus allowing different representations for varying senses of the same word. Specifically, for \emph{N} different sentences where a word \emph{w} is present, ELMo generates \emph{N} different representations of \emph{w} i.e., {$\bm{w}_1, \bm{w}_2, \dot, \bm{w}_N$}.

The mechanism of ELMo is based on the representation obtained from a bidirectional language model. A bidirectional language model (biLM) constitutes of two language models (LM) 1) \textit{forward LM} and 2) \textit{backward LM}. A forward LM takes input representation $\bm{x}_{k}^{LM}$ for each of the $k^{th}$ token and passes it through $L$ layers of forward LSTM to get representations $\overrightarrow { \bm { h } } _ { k , j } ^ { L M }$ where $j=1, \ldots ,L$. Each of these representations, being hidden representations of recurrent neural networks, is context dependent. A forward LM can be seen as a method to model the joint probability of a sequence of tokens:
$p \left( t _ { 1 } , t _ { 2 } , \ldots , t _ { N } \right) = \prod _ { k = 1 } ^ { N } p \left( t _ { k } | t _ { 1 } , t _ { 2 } , \ldots , t _ { k - 1 } \right)$. At a timestep $k-1$ the forward LM predicts the next token $t_{k}$ given the previous observed tokens $t_1, t_2, ..., t_k$. This is typically achieved by placing a softmax layer on top of the final LSTM in a forward LM. On the other hand, a backward LM models the same joint probability of the sequence by predicting the previous token given the future tokens: $p \left( t _ { 1 } , t _ { 2 } , \ldots , t _ { N } \right) = \prod _ { k = 1 } ^ { N } p \left( t _ { k } | t _ { k + 1 } , t _ { k + 2 } , \ldots , t _ { N } \right)$. In other words, a backward LM is similar to forward LM which processes a sequence with the order being reversed. The training of the biLM model involves modeling the log-likelihood of both the sentence orientations. Finally, hidden representations from both LMs are concetenated to compose the final token vectors~\citep{mousa2017contextual}.

For each tokem, ELMo extracts the intermediate layer representations from the biLM and performs a linear combination based on the given downstream task. A $L$-layer biLM contains $2L+1$ set of representations as shown below -
\begin{equation}
\begin{aligned} R _ { k } & = \left\{ \mathbf { x } _ { k } ^ { L M } , \overrightarrow { \mathbf { h } } _ { k , j } ^ { L M } , \overleftarrow { \mathbf { h } } _ { k , j } ^ { L M } | j = 1 , \ldots , L \right\} \\ & = \left\{ \mathbf { h } _ { k , j } ^ { L M } | j = 0 , \ldots , L \right\} \end{aligned}
\end{equation}


Here, $h_{k,0}^{LM}$ is the token representation at the lowest level. One can use either character or word embeddings to initialize $h_{k,0}^{LM}$. For other values of $j$, 
\begin{equation}
h_{k,j}^{LM} = \left[\overrightarrow { \mathbf { h } } _ { k , j } ^ { L M }, \overleftarrow { \mathbf { h } } _ { k , j } ^ { L M }\right] \;\forall j = 1 , \ldots , L. 
\end{equation}
ELMo flattens all layers in $R$ in a single vector such that - 

\begin{equation}
    \text{\textbf{ELMo}} _ { k } ^ { t a s k } = E \left( R _ { k } ; \Theta ^ { t a s k } \right) = \gamma ^ { t a s k } \sum _ { j = 0 } ^ { L } s _ { j } ^ { t a s k } \mathbf { h } _ { k , j } ^ { L M }
    \label{eq:elmo}
\end{equation}
In Eq. \ref{eq:elmo}, $s _ { j } ^ { t a s k }$ is the softmax-normalized weight vector to combine the representations of different layers. $\gamma ^ { t a s k }$ is a hyperparameter which helps in optimization and task specific scaling of the ELMo representation. ELMo produces varied word representations for the same word in different sentences. According to \citet{peters2018deep}, it is always beneficial to combine ELMo word representations with standard global word representations like Glove and Word2Vec. 

Off-late, there has been a surge of interest in pre-trained language models for myriad of natural language tasks~\cite{dai2015semi}. Language modeling is chosen as the pre-training objective as it is widely considered to incorporate multiple traits of natual language understanding and generation. A good language model requires learning complex characteristics of language involving syntactical properties and also semantical coherence. Thus, it is believed that unsupervised training on such objectives would infuse better linguistic knowledge into the networks than random initialization. The \textit{generative pre-training} and \textit{discriminative fine-tuning}
procedure is also desirable as the pre-training is unsupervised and does not require any manual labeling.

~\citet{radford2018improving} proposed similar pre-trained model, the OpenAI-GPT, by adapting the \textit{Transformer} (see section~\ref{sec:transformer}). Recently, \citet{devlin2018bert} proposed BERT which utilizes a transformer network to pre-train a language model for extracting contextual word embeddings. Unlike ELMo and OpenAI-GPT, BERT uses different pre-training tasks for language modeling. In one of the tasks, BERT randomly masks a percentage of words in the sentences and only predicts those masked words. In the other task, BERT predicts the next sentence given a sentence. This task in particular tries to model the relationship among two sentences which is supposedly not captured by traditional bidirectional language models. Consequently, this particular pre-training scheme helps BERT to outperform state-of-the-art techniques by a large margin on key NLP tasks such as QA, Natural Language Inference (NLI) where understanding relation among two sentences is very important. We discuss the impact of these proposed models and the performance achieved by them in section~\ref{sec:contextualembedingsresults}.

The described approaches for contextual word embeddings promises better quality representations for words. The pre-trained deep language models also provide a headstart for downstream tasks in the form of transfer learning. This approach has been extremely popular in computer vision tasks. Whether there would be similar trends in the NLP community, where researchers and practitioners would prefer such models over traditional variants remains to be seen in the future.

\section{Convolutional Neural Networks}\label{sec:3}
Following the popularization of word embeddings and its ability to represent words in a distributed space, the need arose for an effective feature function that extracts higher-level features from constituting words or n-grams. These abstract features would then be used for numerous NLP tasks such as sentiment analysis, summarization, machine translation, and question answering (QA). CNNs turned out to be the natural choice given their effectiveness in computer vision tasks~\cite{krizhevsky2012imagenet, sharif2014cnn, jia2014caffe}.

\begin{figure}[t]
	\includegraphics[scale = 0.5]{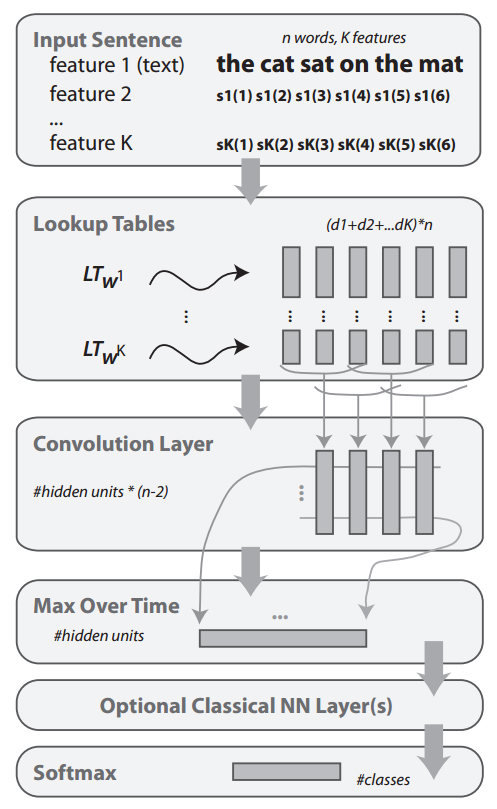}
	\centering
	\caption{CNN framework used to perform word wise class prediction (Figure source:~\citet{collobert2008unified})}\label{fig:collobertCNN}
\end{figure}

The use of CNNs for sentence modeling traces back to~\citet{collobert2008unified}. This work used multi-task learning to output multiple predictions for NLP tasks such as POS tags, chunks, named-entity tags, semantic roles, semantically-similar words and a language model. A look-up table was used to transform each word into a vector of user-defined dimensions. Thus, an input sequence $\{s_1, s_2, ... s_n\}$ of $n$ words was transformed into a series of vectors $\{ {\bm{w}_{s_1}}, {\bm{w}_{s_2}}, ... {\bm{w}_{s_n}} \}$ by applying the look-up table to each of its words (Fig.~\ref{fig:collobertCNN}). 

This can be thought of as a primitive word embedding method whose weights were learned in the training of the network. In~\cite{collobert2011natural}, Collobert extended his work to propose a general CNN-based framework to solve a plethora of NLP tasks. Both these works triggered a huge popularization of CNNs amongst NLP researchers. Given that CNNs had already shown their mettle for computer vision tasks, it was easier for people to believe in their performance. 

CNNs have the ability to extract salient n-gram features from the input sentence to create an informative latent semantic representation of the sentence for downstream tasks. This application was pioneered by ~\citet{collobert2011natural,KalchbrennerACL2014,kim2014convolutional}, which led to a huge proliferation of CNN-based networks in the succeeding literature. Below, we describe the working of a simple CNN-based sentence modeling network: 

\subsection{Basic CNN}

\subsubsection{Sentence Modeling}
\begin{figure}[t]
	\includegraphics[scale=0.3]{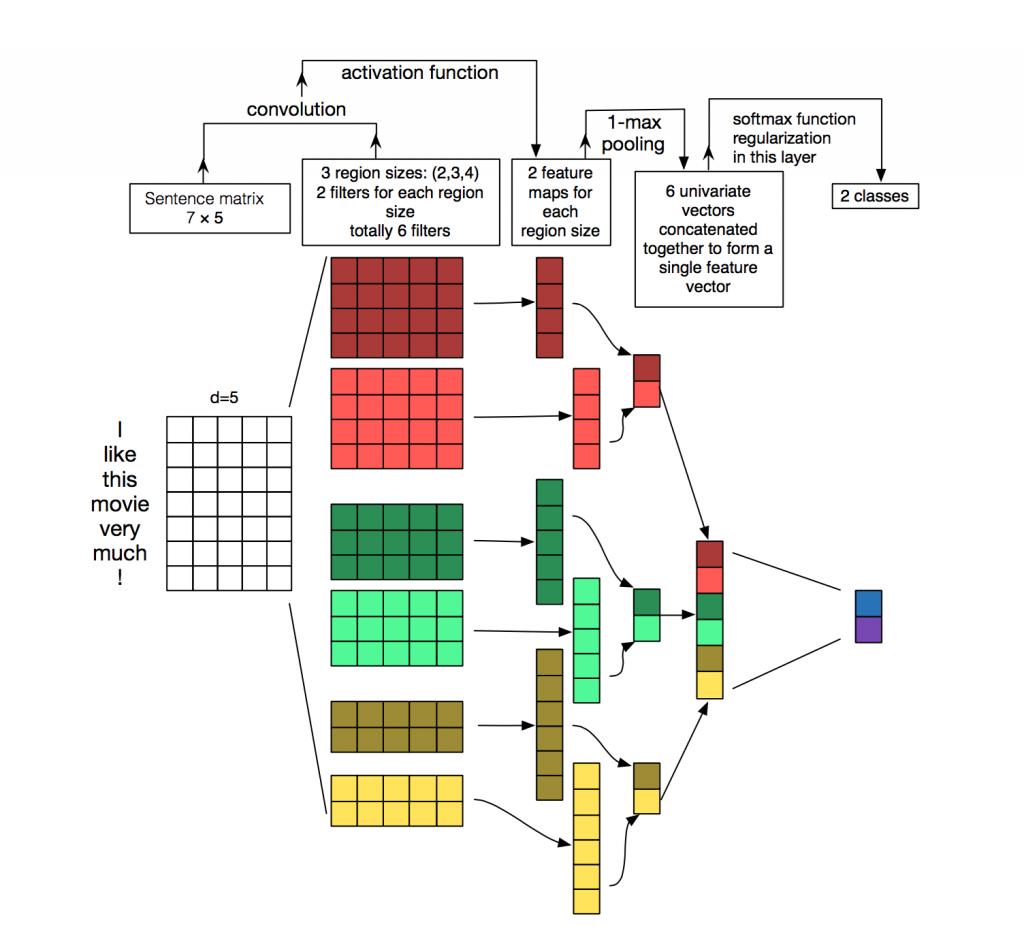}
	\centering
	\caption{CNN modeling on text (Figure source:~\citet{zhang2015sensitivity})}\label{fig:CNN}
\end{figure}

For each sentence, let ${\bf w_{i}} \in \mathcal{R}^d$ represent the word embedding for the $i^{th}$ word in the sentence, where $d$ is the dimension of the word embedding. Given that a sentence has $n$ words, the sentence can now be represented as an embedding matrix ${\bf W} \in \mathcal{R}^{n \times d}$. Fig.~\ref{fig:CNN} depicts such a sentence as an input to the CNN framework.

Let ${\bf w_{i:i+j}} $ refer to the concatenation of vectors $ {\bf w_{i}}, {\bf w_{i+1}}, ... {\bf w_{j}}$. Convolution is performed on this input embedding layer. It involves a \textit{filter} ${\bf k} \in \mathcal{R}^{hd}$ which is applied to a window of $h$ words to produce a new feature. For example, a feature $c_i$ is generated using the window of words ${\bf w_{i:i+h-1}} $ by

\begin{equation}
c_i = f({\bf w_{i:i+h-1}}.{\bf k}^T + b )
\end{equation} 
Here, $b \in \mathcal{R}$ is the bias term and $f$ is a non-linear activation function, for example the hyperbolic tangent. The filter $k$ is applied to all possible windows using the same weights to create the feature map.
\begin{equation}
c = [c_1, c_2, ... , c_{n-h+1}]
\end{equation} 

In a CNN, a number of convolutional filters, also called kernels (typically hundreds), of different widths slide over the entire word embedding matrix. Each kernel extracts a specific pattern of n-gram.
A convolution layer is usually followed by a max-pooling strategy, $\hat{c} = max\{c\}$, which subsamples the input typically by applying a max operation on each filter. This strategy has two primary reasons.

Firstly, max pooling provides a fixed-length output which is generally required for classification. Thus, regardless the size of the filters, max pooling always maps the input to a fixed dimension of outputs. 
Secondly, it reduces the output's dimensionality while keeping the most salient n-gram features across the whole sentence. This is done in a translation invariant manner where each filter is now able to extract a particular feature (e.g., negations) from anywhere in the sentence and add it to the final sentence representation. 

The word embeddings can be initialized randomly or pre-trained on a large unlabeled corpora (as in Section~\ref{sec:2}). The latter option is sometimes found beneficial to performance, especially when the amount of labeled data is limited~\cite{kim2014convolutional}.
This combination of convolution layer followed by max pooling is often stacked to create deep CNN networks. These sequential convolutions help in improved mining of the sentence to grasp a truly abstract representations comprising rich semantic information. The kernels through deeper convolutions cover a larger part of the sentence until finally covering it fully and creating a global summarization of the sentence features.

\subsubsection{Window Approach}
\label{sec:TDNN}
The above-mentioned architecture allows for modeling of complete sentences into sentence representations. However, many NLP tasks, such as NER, POS tagging, and SRL, require word-based predictions. To adapt CNNs for such tasks, a window approach is used, which assumes that the tag of a word primarily depends on its neighboring words. For each word, thus, a fixed-size window surrounding itself is assumed and the sub-sentence ranging within the window is considered. A standalone CNN is applied to this sub-sentence as explained earlier and predictions are attributed to the word in the center of the window. 
Following this approach, \citet{poria2016aspect} employed a multi-level deep CNN to tag each word in a sentence as a possible aspect or non-aspect. Coupled with a set of linguistic patterns, their ensemble classifier managed to perform well in aspect detection. 

The ultimate goal of word-level classification is generally to assign a sequence of labels to the entire sentence. In such cases, structured prediction techniques such as conditional random field (CRF) are sometimes employed to better capture dependencies between adjacent class labels and finally generate cohesive label sequence giving maximum score to the whole sentence~\cite{kirillov2015efficient}.

To get a larger contextual range, the classic window approach is often coupled with a time-delay neural network (TDNN)~\cite{waibel1989phoneme}. Here, convolutions are performed across all windows throughout the sequence. These convolutions are generally constrained by defining a kernel having a certain width. Thus, while the classic window approach only considers the words in the window around the word to be labeled, TDNN considers all windows of words in the sentence at the same time.
At times, TDNN layers are also stacked like CNN architectures to extract local features in lower layers and global features in higher layers~\cite{collobert2011natural}.

\subsection{Applications}
In this section, we present some of the crucial works that employed CNNs on NLP tasks to set state-of-the-art benchmarks in their respective times.

\begin{figure}[t]
   \centering
   \begin{subfigure}[b]{0.45\linewidth}
       \includegraphics[width=\textwidth]{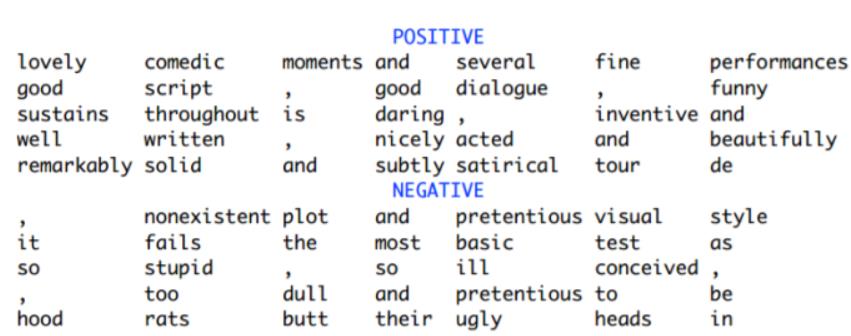}
       \caption{Figure A}
       \label{fig:a}
   \end{subfigure}
   \quad \quad \quad \quad 
   \begin{subfigure}[b]{0.45\textwidth}
       \includegraphics[width=\textwidth]{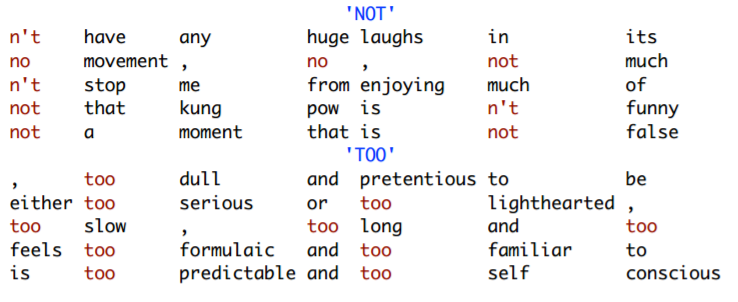}
       \caption{Figure B}
       \label{fig:b}
    \end{subfigure}
    \caption{Top 7-grams by four learned 7-gram kernels; each kernel is sensitive to a specific kind of 7-gram (Figure Source:~\citet{KalchbrennerACL2014})}\label{fig:not-too}
\end{figure}

\citet{kim2014convolutional} explored using the above architecture for a variety of sentence classification tasks, including sentiment, subjectivity and question type classification, showing competitive results. This work was quickly adapted by researchers given its simple yet effective network. After training for a specific task, the randomly initialized convolutional kernels became specific n-gram feature detectors that were useful for that target task (Fig.~\ref{fig:not-too}). This simple network, however, had many shortcomings with the CNN's inability to model long distance dependencies standing as the main issue.

\begin{figure}[t]
	\includegraphics[scale = .45]{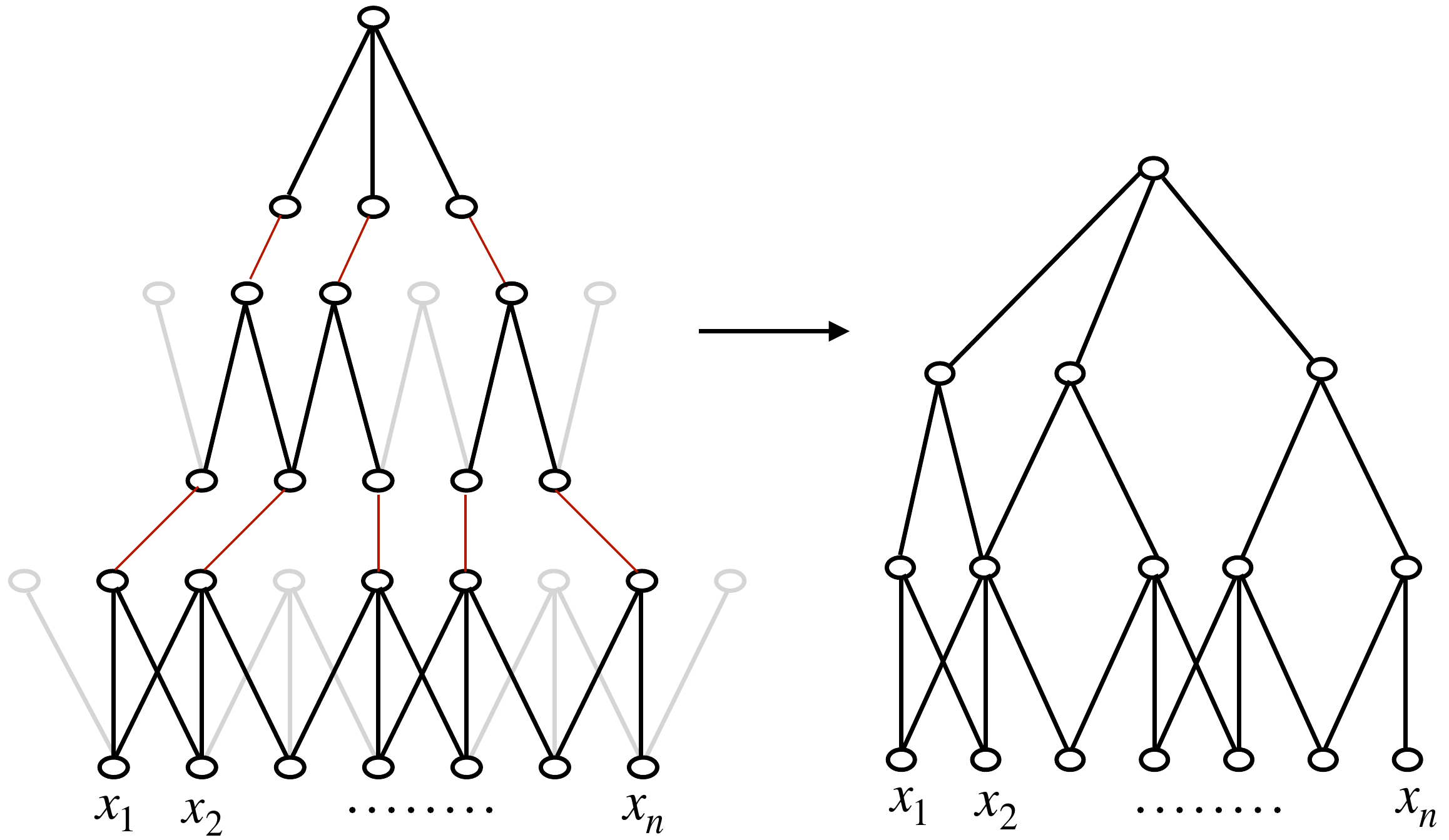}
	\centering
	\caption{DCNN subgraph. With dynamic pooling, a filter with small width at the higher layers can relate
		phrases far apart in the input sentence (Figure Source:~\citet{KalchbrennerACL2014}) }\label{fig:DCNN}
\end{figure}

This issue was partly handled by~\citet{KalchbrennerACL2014}, who published a prominent paper where they proposed a dynamic convolutional neural network (DCNN) for semantic modeling of sentences. They proposed dynamic k-max pooling strategy which, given a sequence ${\bf p}$ selects the $k$ most active features. The selection preserved the order of the features but was insensitive to their specific positions (Fig.~\ref{fig:DCNN}). Built on the concept of TDNN, they added this dynamic k-max pooling strategy to create a sentence model. This combination allowed filters with small width to span across a long range within the input sentence, thus accumulating crucial information across the sentence. In the induced subgraph (Fig.~\ref{fig:DCNN}), higher order features had highly variable ranges that could be either short and focused or global and long as the input sentence. They applied their model on multiple tasks, including sentiment prediction and question type classification, achieving significant results. Overall, this work commented on the range of individual kernels while trying to model contextual semantics and proposed a way to extend their reach.

Tasks involving sentiment analysis also require effective extraction of aspects along with their sentiment polarities~\cite{mukherjee2012aspect}. \citet{ruder2016insight} applied a CNN where in the input they concatenated an aspect vector with the word embeddings to get competitive results. 
CNN modeling approach varies amongst different length of texts. Such differences were seen in many works like~\citet{johnson2015semi}, where performance on longer text worked well as opposed to shorter texts. \citet{wang2015semantic} proposed the usage of CNN for modeling representations of short texts, which suffer from the lack of available context and, thus, require extra efforts to create meaningful representations. The authors proposed semantic clustering which introduced multi-scale semantic units to be used as external knowledge for the short texts. CNN was used to combine these units and form the overall representation. In fact, this requirement of high context information can be thought of as a caveat for CNN-based models. NLP tasks involving microtexts using CNN-based methods often require the need of additional information and external knowledge to perform as per expectations. This fact was also observed in~\cite{porloo}, where authors performed sarcasm detection in Twitter texts using a CNN network. Auxiliary support, in the form of pre-trained networks trained on emotion, sentiment and personality datasets was used to achieve state-of-the-art performance.


CNNs have also been extensively used in other tasks. For example, \citet{denil2014modelling} applied DCNN to map meanings of words that constitute a sentence to that of documents for summarization. The DCNN learned convolution filters at both the sentence and document level, hierarchically learning to capture and compose low-level lexical features into high-level semantic concepts. The focal point of this work was the introduction of a novel visualization technique of the learned representations, which provided insights not only in the learning process but also for automatic summarization of texts. 

CNN models are also suitable for certain NLP tasks that require semantic matching beyond classification~\cite{hu2014convolutional}. A similar model to the above CNN architecture (Fig.~\ref{fig:CNN}) was explored in~\cite{shen2014latent} for information retrieval. The CNN was used for projecting queries and documents to a fixed-dimension semantic space, where cosine similarity between the query and documents was used for ranking documents regarding a specific query. The model attempted to extract rich contextual structures in a query or a document by considering a temporal context window in a word sequence. This captured the contextual features at the word n-gram level. The salient word n-grams is then discovered by the convolution and max-pooling layers which are then aggregated to form the overall sentence vector.

In the domain of QA, \citet{yih2014semantic} proposed to measure the semantic similarity between a question and entries in a knowledge base (KB) to determine what supporting fact in the KB to look for when answering a question. To create semantic representations, a CNN similar to the one in Fig.~\ref{fig:CNN} was used. Unlike the classification setting, the supervision signal came from positive or negative text pairs (e.g., query-document), instead of class labels. Subsequently, \citet{dong2015question} introduced a multi-column CNN (MCCNN) to analyze and understand questions from multiple aspects and create their representations. MCCNN used multiple column networks to extract information from aspects comprising answer types and context from the input questions. By representing entities and relations in the KB with low-dimensional vectors, they used question-answer pairs to train the CNN model so as to rank candidate answers. 
\citet{severyn2016modeling} also used CNN network to model optimal representations of question and answer sentences. They proposed additional features in the embeddings in the form of relational information given by matching words between the question and answer pair. These parameters were tuned by the network. This simple network was able to produce comparable results to state-of-the-art methods.

CNNs are wired in a way to capture the most important information in a sentence. Traditional max-pooling strategies perform this in a translation invariant form. However, this often misses valuable information present in multiple facts within the sentence. To overcome this loss of information for multiple-event modeling, \citet{chen2015event} proposed a modified pooling strategy: dynamic multi-pooling CNN (DMCNN). This strategy used a novel dynamic multi-pooling layer that, as the name suggests, incorporates event triggers and arguments to reserve more crucial information from the pooling layer. 

CNNs inherently provide certain required features like local connectivity, weight sharing, and pooling. This puts forward some degree of invariance which is highly desired in many tasks. Speech recognition also requires such invariance and, thus, \citet{abdel2014convolutional} used a hybrid CNN-HMM model which provided invariance to frequency shifts along the frequency axis. This variability is often found in speech signals due to speaker differences. They also performed limited weight sharing which led to a smaller number of pooling parameters, resulting in lower computational complexity. \citet{palaz2015analysis} performed extensive analysis of CNN-based speech recognition systems when given raw speech as input. They showed the ability of CNNs to directly model the relationship between raw input and phones, creating a robust automatic speech recognition system. 

Tasks like machine translation require perseverance of sequential information and long-term dependency. Thus, structurally they are not well suited for CNN networks, which lack these features. Nevertheless, \citet{tu2015context} addressed this task by considering both the semantic similarity of the translation pair and their respective contexts. Although this method did not address the sequence perseverance problem, it allowed them to get competitive results amongst other benchmarks.

Overall, CNNs are extremely effective in mining semantic clues in contextual windows. However, they are very data heavy models. They include a large number of trainable parameters which require huge training data. This poses a problem when scarcity of data arises. Another persistent issue with CNNs is their inability to model long-distance contextual information and preserving sequential order in their representations~\cite{KalchbrennerACL2014,tu2015context}. Other networks like recursive models (explained below) reveal themselves as better suited for such learning. 

\section{Recurrent Neural Networks}\label{sec:4}
RNNs~\cite{elman1990finding} use the idea of processing sequential information. 
The term ``recurrent" applies as they perform the same task over each instance of the sequence such that the output is dependent on the previous computations and results. Generally, a fixed-size vector is produced to represent a sequence by feeding tokens one by one to a recurrent unit. In a way, RNNs have ``memory" over previous computations and use this information in current processing. This template is naturally suited for many NLP tasks such as language modeling~\cite{mikolov2010recurrent, mikolov2011extensions, sutskever2011generating}, machine translation~\cite{liu2014recursive, auli2013joint, sutskever2014sequence}, speech recognition~\cite{robinson1996use, graves2013speech, graves2014towards, sak2014long}, image captioning~\cite{karpathy2015deep}. This made RNNs increasingly popular for NLP applications in recent years. 

\subsection{Need for Recurrent Networks}
In this section, we analyze the fundamental properties that favored the popularization of RNNs in a multitude of NLP tasks. 
Given that an RNN performs sequential processing by modeling units in sequence, it has the ability to capture the inherent sequential nature present in language, where units are characters, words or even sentences. Words in a language develop their semantical meaning based on the previous words in the sentence. A simple example stating this would be the difference in meaning between ``dog'' and ``hot dog''. RNNs are tailor-made for modeling such context dependencies in language and similar sequence modeling tasks, which resulted to be a strong motivation for researchers to use RNNs over CNNs in these areas. 

Another factor aiding RNN's suitability for sequence modeling tasks lies in its ability to model variable length of text, including very long sentences, paragraphs and even documents~\cite{tang2015document}. Unlike CNNs, RNNs have flexible computational steps that provide better modeling capability and create the possibility to capture unbounded context. This ability to handle input of arbitrary length became one of the selling points of major works using RNNs~\cite{chung2014empirical}.

Many NLP tasks require semantic modeling over the whole sentence. This involves creating a gist of the sentence in a fixed dimensional hyperspace. RNN's ability to summarize sentences led to their increased usage for tasks like machine translation~\cite{cho2014learning} where the whole sentence is summarized to a fixed vector and then mapped back to the variable-length target sequence.

RNN also provides the network support to perform time distributed joint processing. Most of the sequence labeling tasks like POS tagging~\cite{santos2014learning} come under this domain. More specific use cases include applications such as multi-label text categorization~\cite{cheens}, multimodal sentiment analysis~\cite{pordep,tensoremnlp17,traffickingacl17}, and subjectivity detection~\cite{chasub}.

The above points enlist some of the focal reasons that motivated researchers to opt for RNNs. However, it would be gravely wrong to make conclusions on the superiority of RNNs over other deep networks. Recently, several works provided contrasting evidence on the superiority of CNNs over RNNs. Even in RNN-suited tasks like language modeling, CNNs achieved competitive performance over RNNs~\cite{dauphin2016language}.
Both CNNs and RNNs have different objectives when modeling a sentence. While RNNs try to create a composition of an arbitrarily long sentence along with unbounded context, CNNs try to extract the most important n-grams. Although they prove an effective way to capture n-gram features, which is approximately sufficient in certain sentence classification tasks, their sensitivity to word order is restricted locally and long-term dependencies are typically ignored. 

\citet{yin2017comparative} provided interesting insights on the comparative performance between RNNs and CNNs. After testing on multiple NLP tasks that included sentiment classification, QA, and POS tagging, they concluded that there is no clear winner: the performance of each network depends on the global semantics required by the task itself. 

Below, we discuss some of the RNN models extensively used in the literature.

\subsection{RNN models}

\subsubsection{{\bf Simple RNN}}
\begin{figure}[t]
	\includegraphics[width=0.7\linewidth]{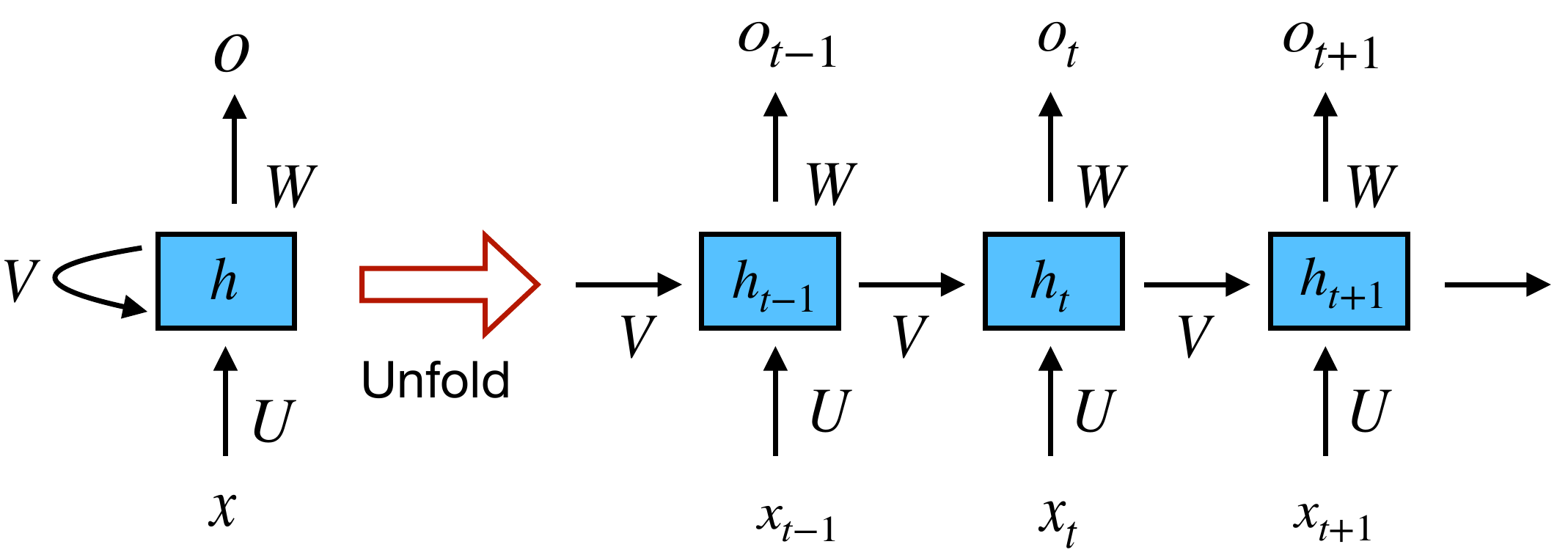}
	\centering
	\caption{Simple RNN network (Figure Source:~\citet{lecun2015deep})}\label{fig:Elman}
\end{figure}

In the context of NLP, RNNs are primarily based on Elman network~\cite{elman1990finding} and they are originally three-layer networks. Fig.~\ref{fig:Elman} illustrates a more general RNN which is unfolded across time to accommodate a whole sequence. 
In the figure, $x_t$ is taken as the input to the network at time step $t$ and $s_t$ represents the hidden state at the same time step. Calculation of $s_t$ is based as per the equation:
\begin{equation}
{\bf s_t} = f( U{\bf x_t} + W{\bf s_{t-1}} )
\end{equation} 
Thus, $s_t$ is calculated based on the current input and the previous time step's hidden state. The function $f$ is taken to be a non-linear transformation such as $tanh, ReLU$ and $U, V, W$ account for weights that are shared across time. In the context of NLP, $x_t$ typically comprises of one-hot encodings or embeddings. At times, they can also be abstract representations of textual content. $o_t$ illustrates the output of the network which is also often subjected to non-linearity, especially when the network contains further layers downstream. 

\begin{figure}[h]
	\includegraphics[width=0.5\linewidth]{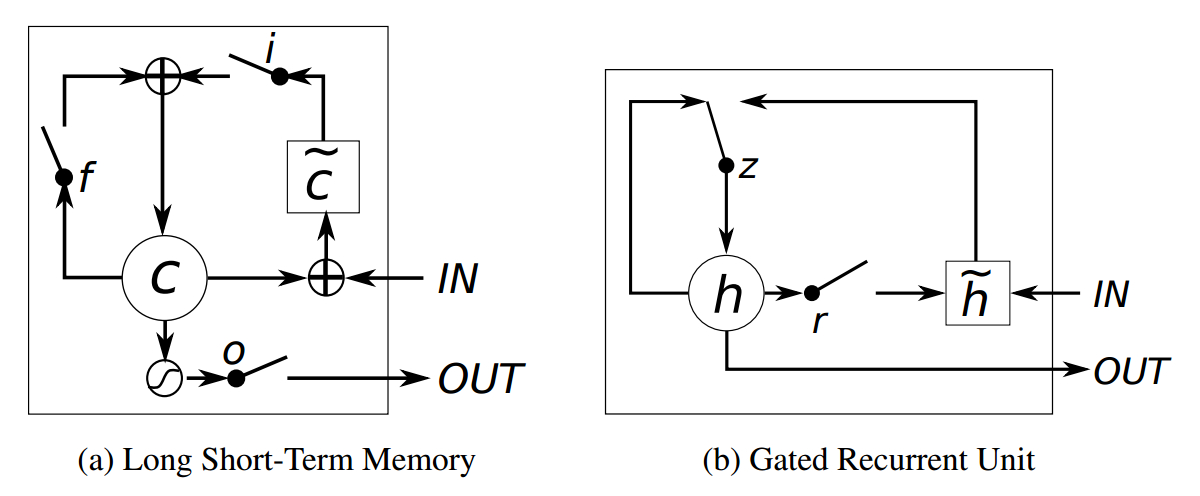}
	\centering
	\caption{Illustration of an LSTM and GRU gate (Figure Source:~\citet{chung2014empirical})}\label{fig:LSTMGRU}
\end{figure}

The hidden state of the RNN is typically considered to be its most crucial element. As stated before, it can be considered as the network's memory element that accumulates information from other time steps. In practice, however, these simple RNN networks suffer from the infamous \emph{vanishing gradient} problem, which makes it really hard to learn and tune the parameters of the earlier layers in the network.

This limitation was overcome by various networks such as long short-term memory (LSTM), gated recurrent units (GRUs), and residual networks (ResNets), where the first two are the most used RNN variants in NLP applications.

\subsubsection{{\bf Long Short-Term Memory}}

LSTM~\cite{hochreiter1997long,gers1999learning} (Fig.~\ref{fig:LSTMGRU}) has additional ``forget" gates over the simple RNN. Its unique mechanism enables it to overcome both the \emph{vanishing} and \emph{exploding} gradient problem.

Unlike the vanilla RNN, LSTM allows the error to back-propagate through unlimited number of time steps. Consisting of three gates: input, forget and output gates, it calculates the hidden state by taking a combination of these three gates as per the equations below: 

\begin{gather}
\small
{\bf x} = \left[ {\begin{array}{c}
	\bm{h}_{t-1}\\
	\bm{x}_t \\
	\end{array} } \right]\\
f_t = \sigma(W_f . {\bf x} + b_f)\\
i_t = \sigma(W_i . {\bf x} + b_i) \\
o_t = \sigma(W_o . {\bf x} + b_o)\\
c_t = f_t \odot c_{t-1} + i_t \odot tanh(W_c . X + b_c) \\
h_t = o_t \odot tanh(c_t)
\end{gather}

\subsubsection{{\bf Gated Recurrent Units}}
Another gated RNN variant called GRU~\cite{cho2014learning} (Fig.~\ref{fig:LSTMGRU}) of lesser complexity was invented with empirically similar performances to LSTM in most tasks. GRU comprises of two gates, reset gate and update gate, and handles the flow of information like an LSTM sans a memory unit. Thus, it exposes the whole hidden content without any control. Being less complex, GRU can be a more efficient RNN than LSTM. The working of GRU is as follows:
\begin{gather}
{\bf z} = \sigma(U_z . {\bf x_t} + W_z . {\bf h_{t-1}})\\
{\bf r} = \sigma(U_r . {\bf x_t} + W_r . {\bf h_{t-1}})\\
{\bf s_t} = tanh(U_z . {\bf x_t} + W_s .( h_{t-1} \odot {\bf r}) )\\
{\bf h_t} = (1 - {\bf z}) \odot {\bf s_t} + {\bf z} \odot {\bf h_{t-1}} 
\end{gather}

\begin{figure}[ht]
	\includegraphics[width=0.6\linewidth]{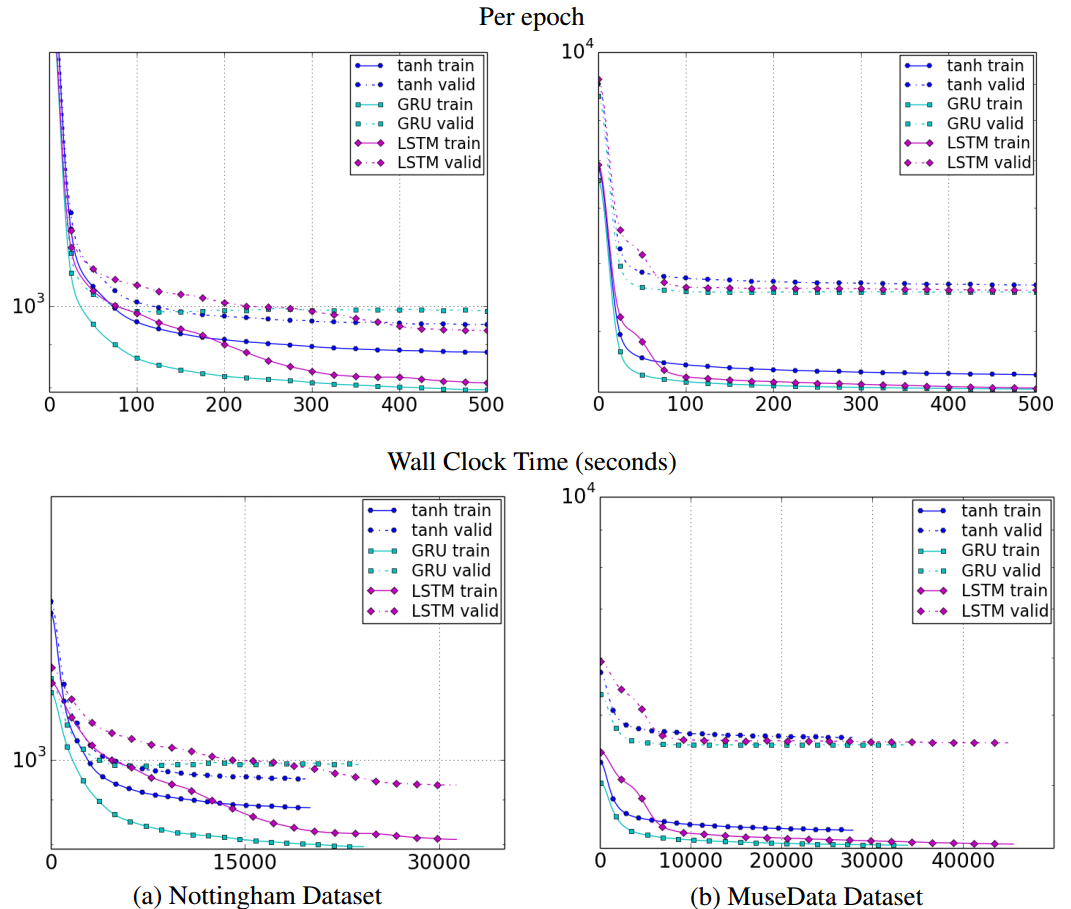}
	\centering
	\caption{Learning curves for training and validation sets of different types of units with respect to (top) the number of iterations and (bottom) the wall clock time. y-axis corresponds to the negative log	likelihood of the model shown in log-scale (Figure source:~\citet{chung2014empirical})}\label{fig:LSTMvsGRU}
\end{figure}

Researchers often face the dilemma of choosing the appropriate gated RNN. This also extends to developers working in NLP. Throughout the history, most of the choices over the RNN variant tended to be heuristic.
\citet{chung2014empirical} did a critical comparative evaluation of the three RNN variants mentioned above, although not on NLP tasks. They evaluated their work on tasks relating to polyphonic music modeling and speech signal modeling. Their evaluation clearly demonstrated the superiority of the gated units (LSTM and GRU)
over the traditional simple RNN (in their case, using $tanh$ activation) (Fig.~\ref{fig:LSTMvsGRU}).
However, they could not make any concrete conclusion about which of the two gating units was better. This fact has been noted in other works too and, thus, people often leverage on other factors like computing power while choosing between the two.

\subsection{Applications}

\subsubsection{RNN for word-level classification}
RNNs have had a huge presence in the field of word-level classification. Many of their applications stand as state of the art in their respective tasks. 
\citet{lample2016neural} proposed to use bidirectional LSTM for NER. The network captured arbitrarily long context information around the target word (curbing the limitation of a fixed window size) resulting in two fixed-size vector, on top of which another fully-connected layer was built. They used a CRF layer at last for the final entity tagging. 

RNNs have also shown considerable improvement in language modeling over traditional methods based on count statistics. Pioneering work in this field was done by~\citet{graves2013generating}, who introduced the effectiveness of RNNs in modeling complex sequences with long range context structures. He also proposed deep RNNs where multiple layers of hidden states were used to enhance the modeling. This work established the usage of RNNs on tasks beyond the context of NLP. Later, \citet{sundermeyer2015feedforward} compared the gain obtained by replacing a feed-forward neural network with an RNN when conditioning the prediction of a word on the words ahead. In their work, they proposed a typical hierarchy in neural network architectures where feed-forward neural networks gave considerable improvement over traditional count-based language models, which in turn were superseded by RNNs and later by LSTMs. An important point that they mentioned was the applicability of their conclusions to a variety of other tasks such as statistical machine translation~\cite{sundermeyer2014translation}.

\subsubsection{RNN for sentence-level classification}
\citet{wang2015predicting} proposed encoding entire tweets with LSTM, whose hidden state is used for predicting sentiment polarity. This simple strategy proved competitive to the more complex DCNN structure by~\citet{KalchbrennerACL2014} designed to endow CNN models with ability to capture long-term dependencies. In a special case studying negation phrase, the authors also showed that the dynamics of LSTM gates can capture the reversal effect of the word \emph{not}. 

Similar to CNN, the hidden state of an RNN can also be used for semantic matching between texts. In dialogue systems, \citet{lowe2015ubuntu} proposed to match a message with candidate responses with Dual-LSTM, which encodes both as fixed-size vectors and then measure their inner product as the basis to rank candidate responses. 

\subsubsection{RNN for generating language}
A challenging task in NLP is generating natural language, which is another natural application of RNNs. Conditioned on textual or visual data, deep LSTMs have been shown to generate reasonable task-specific text in tasks such as machine translation, image captioning, etc. In such cases, the RNN is termed a decoder. 

In~\cite{sutskever2014sequence}, the authors proposed a general deep LSTM encoder-decoder framework that maps a sequence to another sequence. One LSTM is used to encode the ``source'' sequence as a fixed-size vector, which can be text in the original language (machine translation), the question to be answered (QA) or the message to be replied to (dialogue systems). The vector is used as the initial state of another LSTM, named the decoder. During inference, the decoder generates tokens one by one, while updating its hidden state with the last generated token. Beam search is often used to approximate the optimal sequence. 

\begin{figure}[t]
	\includegraphics[width=0.5\linewidth]{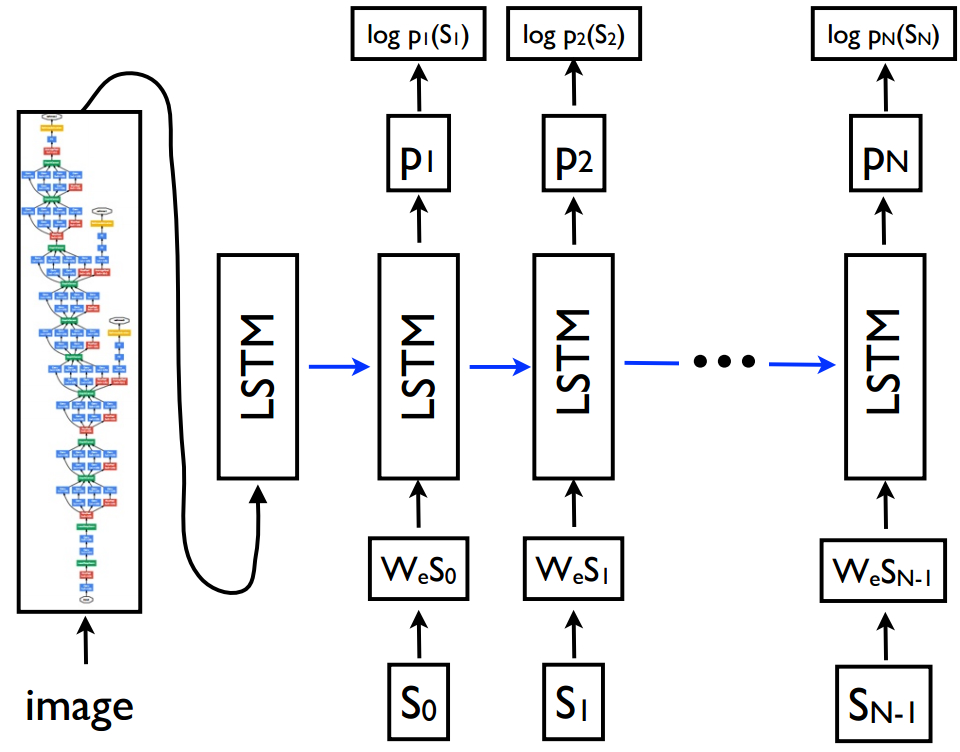}
	\centering
	\caption{LSTM decoder combined with a CNN image embedder to generate image captioning (Figure source:~\citet{vinyals2015show})}\label{fig:cnn_lstm_image_captioning}
\end{figure}

\citet{sutskever2014sequence} experimented with 4-layer LSTM on a machine translation task in an end-to-end fashion, showing competitive results. In~\cite{vinyals2015neural}, the same encoder-decoder framework is employed to model human conversations. When trained on more than 100 million message-response pairs, the LSTM decoder is able to generate very interesting responses in the open domain. It is also common to condition the LSTM decoder on additional signal to achieve certain effects. In~\cite{li2016persona}, the authors proposed to condition the decoder on a constant persona vector that captures the personal information of an individual speaker. 
In the above cases, language is generated based mainly on the semantic vector representing textual input. Similar frameworks have also been successfully used in image-based language generation, where visual features are used to condition the LSTM decoder (Fig.~\ref{fig:cnn_lstm_image_captioning}). 

\begin{figure}[b]
	\includegraphics[width=0.7\linewidth]{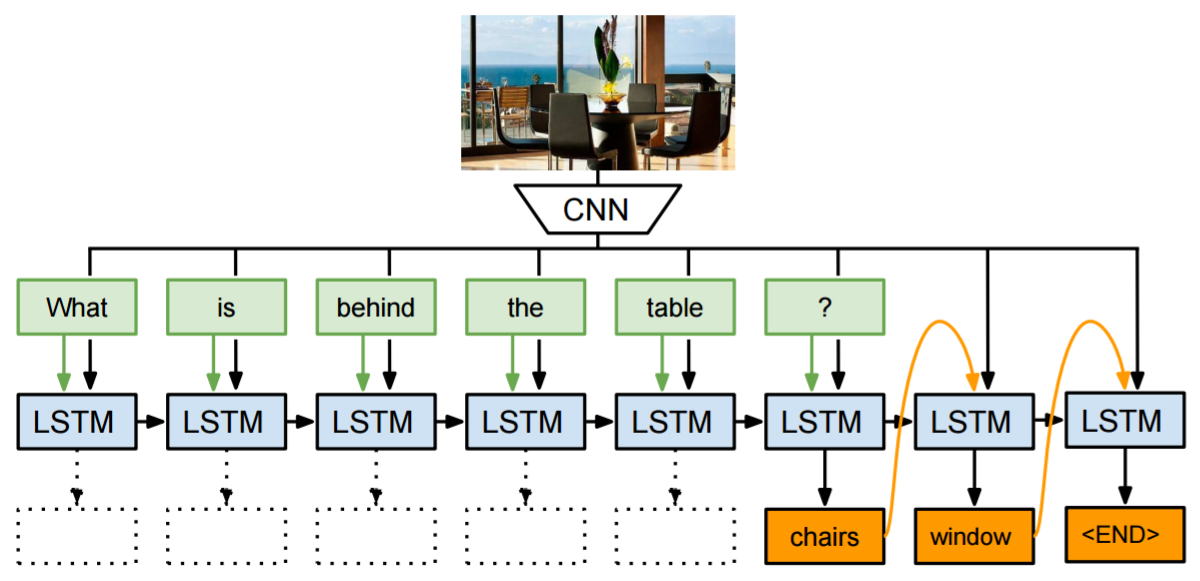}
	\centering
	\caption{Neural-image QA (Figure source:~\citet{malinowski2015ask})}\label{fig:Neural-image-QA}
\end{figure}

Visual QA is another task that requires language generation based on both textual and visual clues. \citet{malinowski2015ask} were the first to provide an end-to-end deep learning solution where they predicted the answer as a set of words conditioned on the input image modeled by a CNN and text modeled by an LSTM (Fig.~\ref{fig:Neural-image-QA}). 

\citet{kumar2015ask} tackled this problem by proposing an elaborated network termed dynamic memory network (DMN), which had four sub-modules. The idea was to repeatedly attend to the input text and image to form episodes of information improved at each iteration. Attention networks were used for fine-grained focus on input text phrases.

\subsection{Attention Mechanism}
One potential problem that the traditional encoder-decoder framework faces is that the encoder at times is forced to encode information which might not be fully relevant to the task at hand. The problem arises also if the input is long or very information-rich and selective encoding is not possible. 

For example, the task of text summarization can be cast as a sequence-to-sequence learning problem, where the input is the original text and the output is the condensed version. Intuitively, it is unrealistic to expect a fixed-size vector to encode all information in a piece of text whose length can potentially be very long. Similar problems have also been reported in machine translation~\cite{bahdanau2014neural}.

In tasks such as text summarization and machine translation, certain alignment exists between the input text and the output text, which means that each token generation step is highly related to a certain part of the input text. This intuition inspires the attention mechanism. This mechanism attempts to ease the above problems by allowing the decoder to refer back to the input sequence. Specifically during decoding, in addition to the last hidden state and generated token, the decoder is also conditioned on a ``context'' vector calculated based on the input hidden state sequence. The attention mechanism can be broadly seen as mapping a query and a set of key-value pairs to an output, where all the mentioned components are vectors. The output is a combination of the values whose weights are determined by the compatibility between the query and the corresponding keys. This output amounts to the ``context'' of the input used in decoding the output.

\begin{wrapfigure}{R}{0.5\textwidth}
\centering
\includegraphics[scale=0.4]{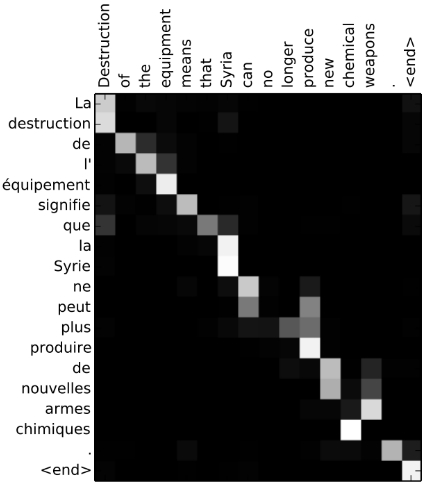}
	\centering
	\caption{Word alignment matrix (Figure source:~\citet{bahdanau2014neural}) }\label{fig:word-alignment-matrix}
\end{wrapfigure}

\citet{bahdanau2014neural} first applied the attention mechanism to machine translation, which improved the performance especially for long sequences. In their work, the attention signal over the input hidden state sequence is determined with a multi-layer perceptron by the last hidden state of the decoder. By visualizing the attention signal over the input sequence during each decoding step, a clear alignment between the source and target language can be demonstrated (Fig.~\ref{fig:word-alignment-matrix}). 

A similar approach was applied to the task of summarization by~\citet{rush2015neural} where each output word in the summary was conditioned on the input sentence through an attention mechanism. The authors performed abstractive summarization which is not very conventional as opposed to extractive summarization, but can be scaled up to large data with minimal linguistic input.

In image captioning, \citet{xu2015show} conditioned the LSTM decoder on different parts of the input image during each decoding step. Attention signal was determined by the previous hidden state and CNN features. In~\cite{vinyals2015grammar}, the authors casted the syntactical parsing problem as a sequence-to-sequence learning task by linearizing the parsing tree. The attention mechanism proved to be more data-efficient in this work. A further step in referring to the input sequence was to directly copy words or sub-sequences of the input onto the output sequence under a certain condition~\cite{vinyals2015pointer}, which was useful in tasks such as dialogue generation and text summarization. Copying or generation was chosen at each time step during decoding~\cite{paulus2017deep}.

\begin{figure}[t]
	\includegraphics[width=0.4\linewidth]{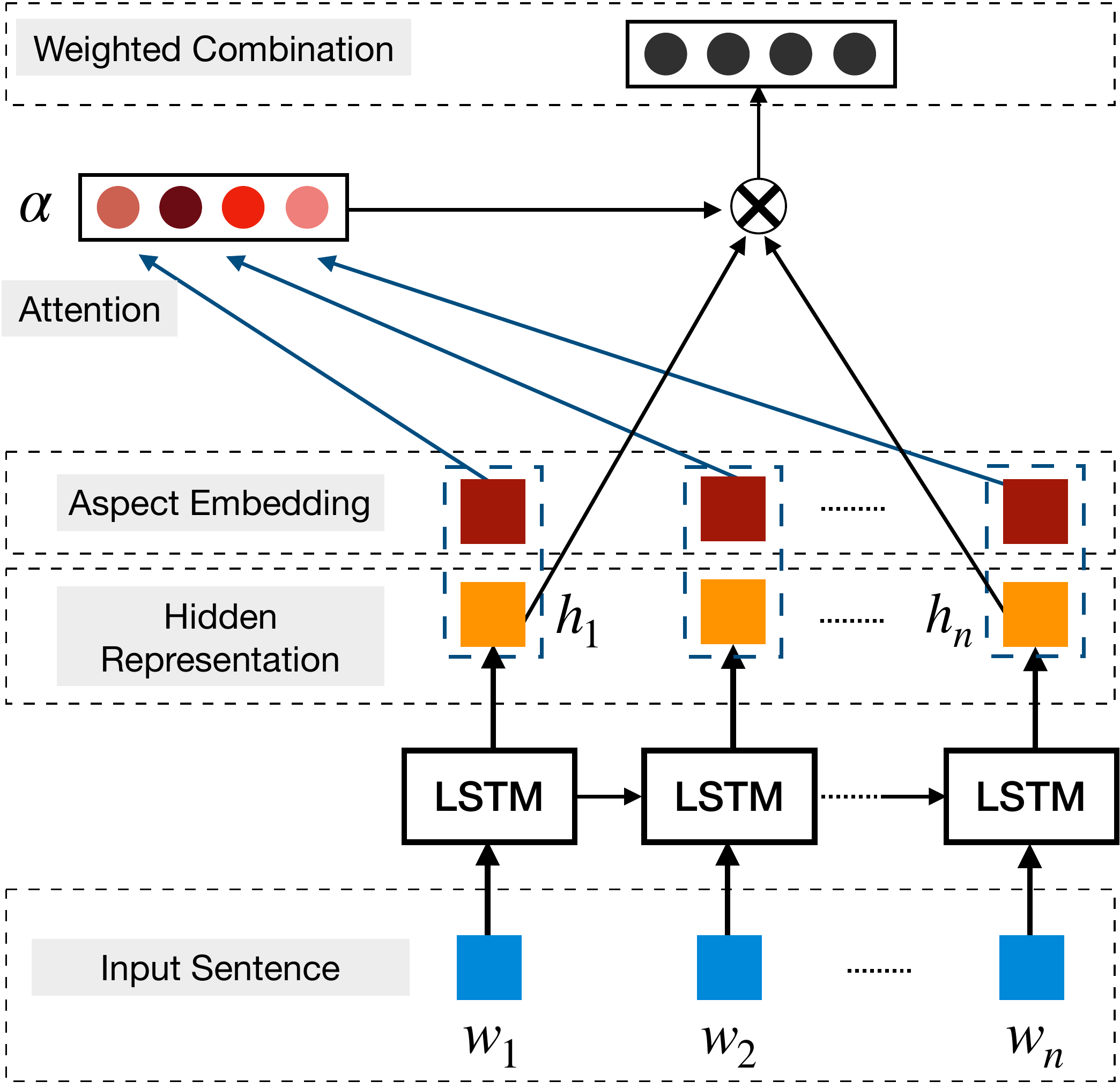}
	\centering
	\caption{Aspect classification using attention (Figure source: ~\citet{wang2016attention}) }\label{fig:aspectattention}
\end{figure}

\begin{figure}[b]
	\includegraphics[width=0.4\linewidth]{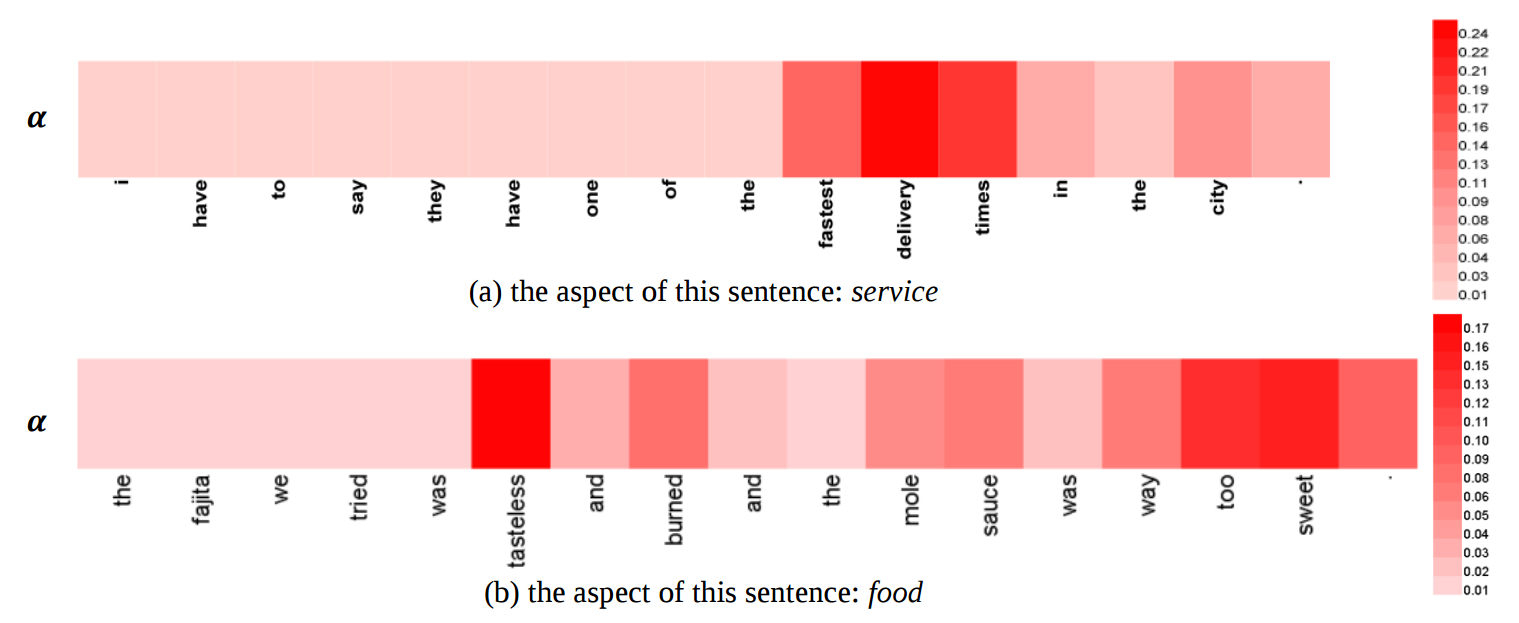}
	\centering
	\caption{Focus of attention module on the sentence for certain aspects (Figure source:~\citet{wang2016attention}) }\label{fig:heatmap}
	\label{fig:multiheadattention}
\end{figure}

In aspect-based sentiment analysis, \citet{wang2016attention} proposed an attention-based solution where they used aspect embeddings to provide additional support during classification (Fig.~\ref{fig:aspectattention}). The attention module focused on selective regions of the sentence which affected the aspect to be classified. 
This can be seen in Fig.~\ref{fig:heatmap} where, for the aspect \emph{service} in (a), the attention module dynamically focused on the phrase ``fastest delivery times"
and in (b) with the aspect \emph{food}, it identified multiple key-points across the sentence that included ``tasteless'' and ``too sweet''. Recently, \citet{ma2018targeted} augmented LSTM with a hierarchical attention mechanism consisting of a target-level attention and a sentence-level attention to exploit commonsense knowledge for targeted aspect-based sentiment analysis.

On the other hand, \citet{tang2016aspect} adopted a solution based on a memory network (also known as MemNet~\cite{weston2014memory}), which employed multiple-hop attention. The multiple attention computation layer on the memory led to improved lookup for most informational regions in the memory and subsequently aided the classification. Their work stands as the state of the art in this domain.

\begin{figure}[t]
	\includegraphics[width=0.3\linewidth]{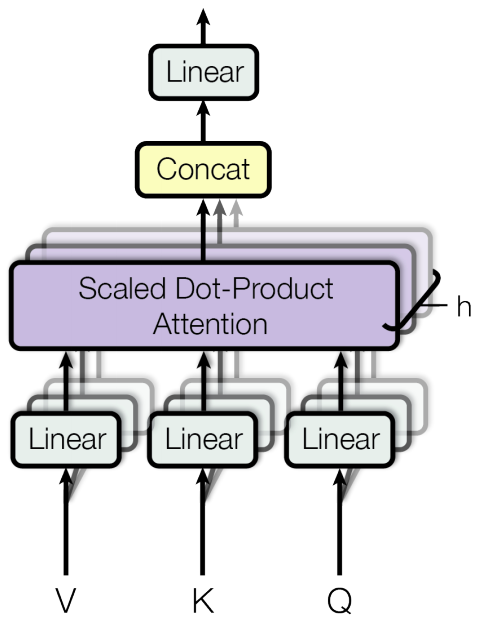}
	\centering
	\caption{Multi-head Attention:~\citet{Vaswani2017attention}) }\label{fig:multi-head-attention}
\end{figure}

Given the intuitive applicability of attention modules, they are still being actively investigated by NLP researchers and adopted for an increasing number of applications. 

\subsection{Parallelized Attention: The Transformer} \label{sec:transformer}

Both CNNs and RNNs have been crucial in sequence transduction applications involving the encoder-decoder architecture. Attention-based mechanisms, as described above, have further boosted the capabilities of these models. However, one of the bottlenecks suffered by these architectures is the sequential processing at the encoding step. To address this, ~\citet{Vaswani2017attention} proposed the \textit{Transformer} which dispensed the recurrence and convolutions involved in the encoding step entirely and based models only on attention mechanisms to capture the global relations between input and output. As a result, the overall architecture became more parallelizable and required lesser time to train along with positive results on tasks ranging from translation to parsing.

The Transformer consists stacked layers in both encoder and decoder components. Each layer has two sub-layers comprising \textit{multi-head attention layer} (Figure~\ref{fig:multiheadattention}) followed by a position-wise feed forward network. For set of queries $Q$, keys $K$ and values $V$, the multi-head attention module performs attention $h$ times where the computation can be seen as:
\begin{align}
    \text{MultiHead}(Q,K,V) = \text{Concat}(\text{head}_1, \text{head}_2, \ldots, \text{head}_h) W^o \\
    \text{where}\;\text{head}_i = \text{Attention}(QW_i^Q, KW_i^K, VW_i^V)\\
    \text{and}\;\text{Attention}(Q,K,V)= \text{softmax}\left(\frac{QK^T}{\sqrt{d_k}}\right)V
\end{align}
here, $W_i^{[.]}$ and $W^o$ are projection parameters. Incorporating other techniques such as residual connections~\cite{he2016deep}, layer normalization~\cite{ba2016layer}, dropouts, positional encodings, and others, the model achieves state-of-the-art results in English-German and English-French translation and constituency parsing.

\section{Recursive Neural Networks}\label{sec:5}
Recurrent neural networks represent a natural way to model sequences. Arguably, however, language exhibits a natural recursive structure, where words and sub-phrases combine into phrases in a hierarchical manner. Such structure can be represented by a constituency parsing tree. Thus, tree-structured models have been used to better make use of such syntactic interpretations of sentence structure~\cite{socher2013recursive}. Specifically, in a recursive neural network, the representation of each non-terminal node in a parsing tree is determined by the representations of all its children.

\subsection{Basic model}
\begin{figure}
   \centering
   \begin{subfigure}[b]{0.4\linewidth}
       \includegraphics[width=\textwidth]{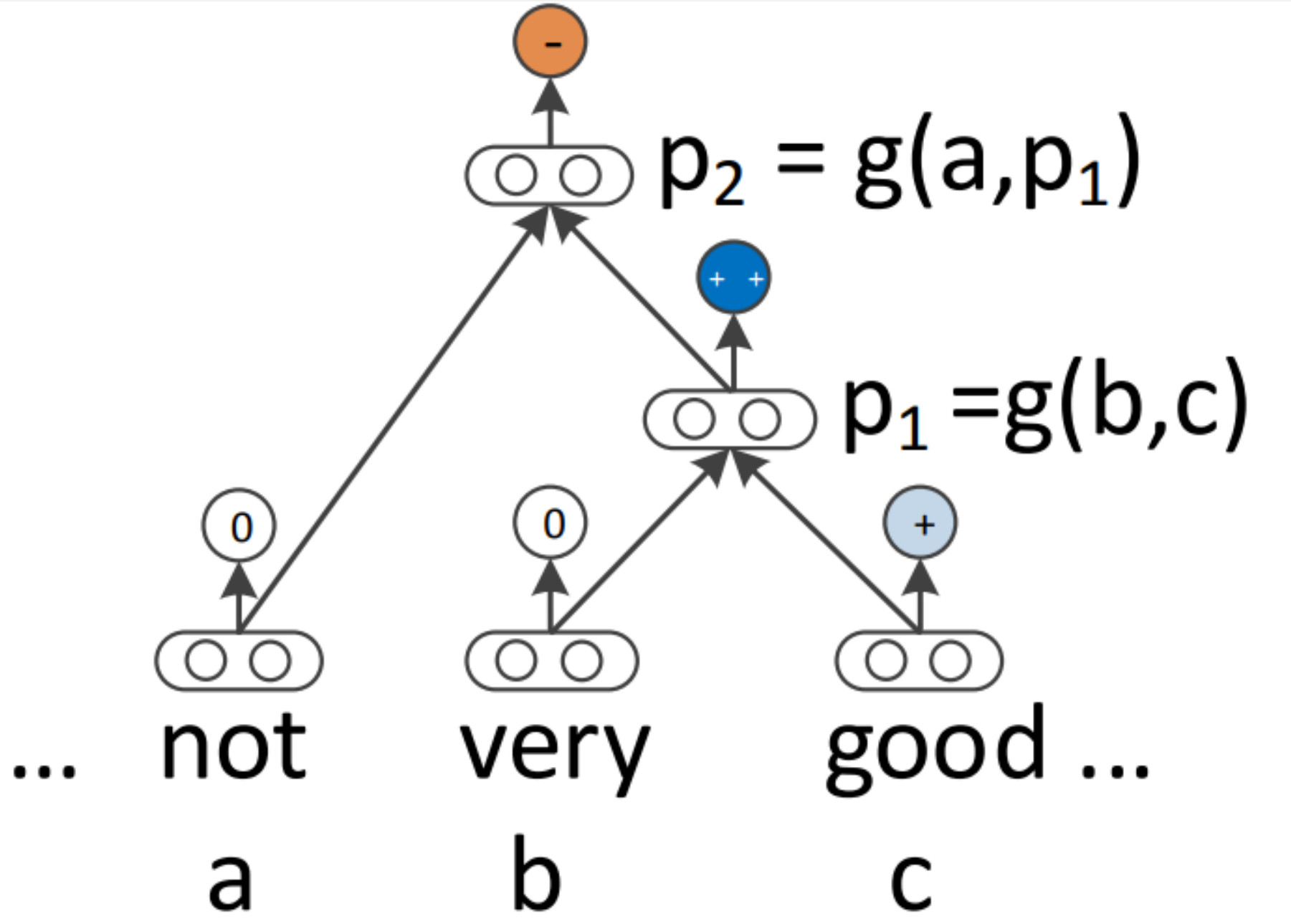}
       \caption{Recursive neural networks for phrase-level sentiment classification (Figure source:~\citet{socher2013recursive})}
       \label{fig:recursiveNN}
   \end{subfigure}
   \quad \quad \quad
   \begin{subfigure}[b]{0.4\textwidth}
       \includegraphics[width=\textwidth]{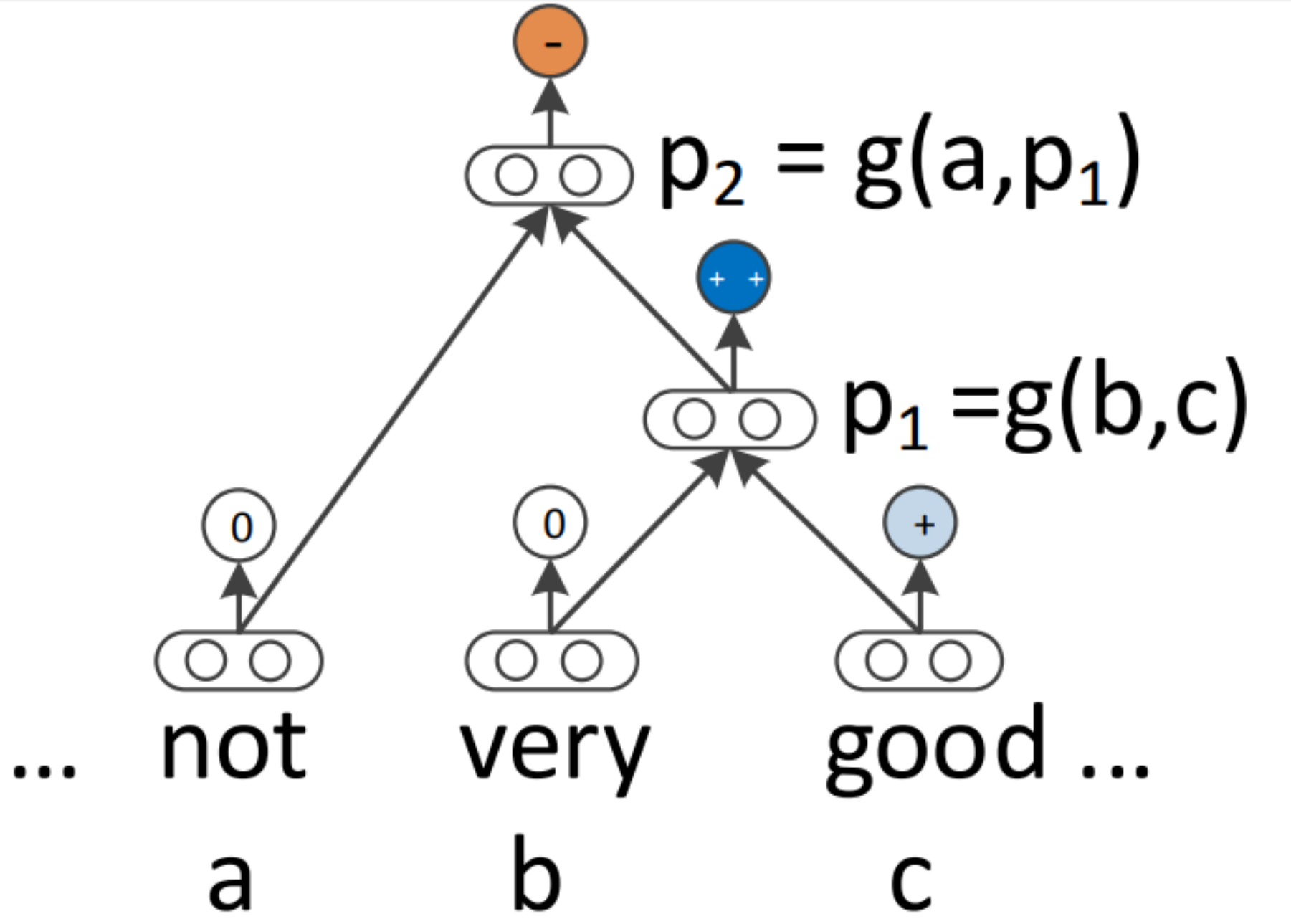}
       \caption{Recursive neural networks iteratively form high-level representation from lower-level representations.}
       \label{fig:recursiveNN2}
    \end{subfigure}
    \caption{Recursive Neural Networks}
\end{figure}

In this section, we describe the basic structure of recursive neural networks. As shown in Fig.~\ref{fig:recursiveNN} and~\ref{fig:recursiveNN2}, the network $g$ defines a compositional function on the representations of phrases or words ($b, c$ or $a, p_1$) to compute the representation of a higher-level phrase ($p_1$ or $p_2$). The representations of all nodes take the same form.

\citet{socher2013recursive} described multiple variations of this model. In its simplest form, $g$ is defined as: 
\begin{gather}
\small
{p_1} = tanh\left(W\left[ {\begin{array}{c}
	b\\
	c \\
	\end{array} } \right]\right), {p_2} = tanh\left(W\left[ {\begin{array}{c}
	a\\
	p_1 \\
	\end{array} } \right]\right)
\end{gather}
in which the representation for each node is a $d$-dimensional vector and $W\in\mathcal{R}^{D\times{2D}}$.

Another variation is the MV-RNN~\cite{socher2012semantic}. The idea is to represent every word and phrase as both a matrix and a vector. When two constituents are combined, the matrix of one is multiplied with the vector of the other:
\begin{gather}
\small
{p_1} = tanh\left(W\left[ {\begin{array}{c}
	Cb\\
	Bc \\
	\end{array} } \right]\right), {P_1} = tanh\left(W_M\left[ {\begin{array}{c}
	B\\
	C \\
	\end{array} } \right]\right)
\end{gather}
in which $b,c,p_1\in\mathcal{R}^{D}$, $B,C,P_1\in\mathcal{R}^{D\times{D}}$, and $W_M\in\mathcal{R}^{D\times{2D}}$. Compared to the vanilla form, MV-RNN parameterizes the compositional function with matrices corresponding to the constituents.

The recursive neural tensor network (RNTN) is proposed to introduce more interaction between the input vectors without making the number of parameters exceptionally large like MV-RNN. RNTN is defined by:
\begin{gather}
\small
{p_1} = tanh\left(\left[ {\begin{array}{c}
	b\\
	c\\
	\end{array} } \right]^{T}V^{[1:D]}\left[ {\begin{array}{c}
	b\\
	c\\
	\end{array} } \right]+
  W\left[ {\begin{array}{c}
	b\\
	c\\
	\end{array} } \right]\right)
\end{gather}
where $V\in\mathcal{R}^{2D\times{2D}\times{D}}$ is a tensor that defines multiple bilinear forms.



\subsection{Applications}
One natural application of recursive neural networks is parsing~\cite{socher2011parsing}. A scoring function is defined on the phrase representation to calculate the plausibility of that phrase. Beam search is usually applied for searching the best tree. The model is trained with the max-margin objective~\cite{taskar2004max}.

\begin{figure}[t]
	\includegraphics[scale=0.6]{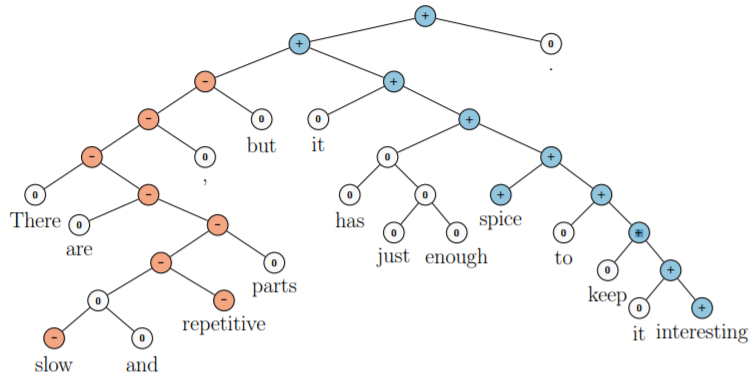}
	\centering
	\caption{Recursive neural networks applied on a sentence for sentiment classification. Note that ``but'' plays a crucial role on determining the sentiment of the whole sentence (Figure source:~\citet{socher2013recursive})}\label{fig:but}
\end{figure}

Based on recursive neural networks and the parsing tree, \citet{socher2013recursive} proposed a phrase-level sentiment analysis framework (Fig.~\ref{fig:but}), where each node in the parsing tree can be assigned a sentiment label.

\citet{socher2012semantic} classified semantic relationships such as cause-effect or topic-message between nominals in a sentence by building a single compositional semantics for the minimal constituent including both terms. \citet{bowman2014recursive} proposed to classify the logical relationship between sentences with recursive neural networks. The representations for both sentences are fed to another neural network for relationship classification. They show that both vanilla and tensor versions of the recursive unit performed competitively in a textual entailment dataset.

To avoid the gradient vanishing problem, LSTM units have also been applied to tree structures in~\cite{tai2015improved}. The authors showed improved sentence representation over linear LSTM models, as clear improvement in sentiment analysis and sentence relatedness test was observed.

\section{Deep reinforced models and deep unsupervised learning}\label{sec:6}

\subsection{Reinforcement learning for sequence generation}
Reinforcement learning is a method of training an agent to perform discrete actions before obtaining a reward. In NLP, tasks concerning language generation can sometimes be cast as reinforcement learning problems.

In its original formulation, RNN language generators are typically trained by maximizing the likelihood of each token in the ground-truth sequence given the current hidden state and the previous tokens. Termed ``teacher forcing'', this training scheme provides the real sequence prefix to the generator during each generation (loss evaluation) step. At test time, however, ground-truth tokens are then replaced by a token generated by the model itself. This discrepancy between training and inference, termed ``exposure bias''~\cite{bengio2015scheduled, ranzato2015sequence}, can yield errors that can accumulate quickly along the generated sequence.

Another problem with the word-level maximum likelihood strategy, when training auto-regressive language generation models, is that the training objective is different from the test metric. It is unclear how the n-gram overlap based metrics (BLEU, ROUGE) used to evaluate these tasks (machine translation, dialogue systems, etc.) can be optimized with the word-level training strategy. Empirically, dialogue systems trained with word-level maximum likelihood also tend to produce dull and short-sighted responses~\cite{li2016deep}, while text summarization tends to produce incoherent or repetitive summaries~\cite{paulus2017deep}. 

Reinforcement learning offers a prospective to solve the above problems to a certain extent. In order to optimize the non-differentiable evaluation metrics directly, \citet{ranzato2015sequence} applied the REINFORCE algorithm~\cite{williams1992simple} to train RNN-based models for several sequence generation tasks (e.g., text summarization, machine translation and image captioning), leading to improvements compared to previous supervised learning methods. In such a framework, the generative model (RNN) is viewed as an agent, which interacts with the external environment (the words and the context vector it sees as input at every time step). 
The parameters of this agent defines a policy, whose execution results in the agent picking an action, which refers to predicting the next word in the sequence at each time step. After taking an action the agent updates its internal state (the hidden units of RNN). Once the agent has reached the end of a sequence, it observes a reward. This reward can be any developer-defined metric tailored to a specific task. For example, \citet{li2016deep} defined 3 rewards for a generated sentence based on ease of answering, information flow, and semantic coherence. 

There are two well-known shortcomings of reinforcement learning. To make reinforcement learning tractable, it is desired to carefully handle the state and action space~\cite{young2010hidden,young2013pomdp}, which in the end may restrict expressive power and learning capacity of the model. Secondly, the need for training the reward functions makes such models hard to design and measure at run time~\cite{su2015learning,su2016line}.

Another approach for sequence-level supervision is to use the adversarial training technique~\cite{goodfellow2014generative}, where the training objective for the language generator is to fool another discriminator trained to distinguish generated sequences from real sequences. The generator \emph{G} and the discriminator \emph{D} are trained jointly in a min-max game which ideally leads to \emph{G}, generating sequences indistinguishable from real ones. This approach can be seen as a variation of generative adversarial networks in~\cite{goodfellow2014generative}, where \emph{G} and \emph{D} are conditioned on certain stimuli (for example, the source image in the task of image captioning). In practice, the above scheme can be realized under the reinforcement learning paradigm with policy gradient. For dialogue systems, the discriminator is analogous to a human Turing tester, who discriminates between human and machine-produced dialogues~\cite{li2017adversarial}.
 
\subsection{Unsupervised sentence representation learning}
Similar to word embeddings, distributed representation for sentences can also be learned in an unsupervised fashion. The result of such unsupervised learning are ``sentence encoders'', which map arbitrary sentences to fixed-size vectors that can capture their semantic and syntactic properties. Usually an auxiliary task has to be defined for the learning process.

Similar to the skip-gram model~\cite{mikolov2013efficient} for learning word embeddings, the skip-thought model~\cite{kiros2015skip} was proposed for learning sentence representation, where the auxiliary task was to predict two adjacent sentences (before and after) based on the given sentence. The seq2seq model was employed for this learning task. One LSTM encoded the sentence to a vector (distributed representation). Two other LSTMs decoded such representation to generate the target sequences. Standard seq2seq training process was used. After training, the encoder could be seen as a generic feature extractor (word embeddings were also learned in the same time). 

\citet{kiros2015skip} verified the quality of the learned sentence encoder on a range of sentence classification tasks, showing competitive results with a simple linear model based on the static feature vectors. However, the sentence encoder can also be fine-tuned in the supervised learning task as part of the classifier.
\citet{dai2015semi} investigated the use of the decoder to reconstruct the encoded sentence itself, which resembled an autoencoder~\cite{rumelhart1985learning}. 

Language modeling could also be used as an auxiliary task when training LSTM encoders, where the supervision signal came from the prediction of the next token. \citet{dai2015semi} conducted experiments on initializing LSTM models with learned parameters on a variety of tasks. They showed that pre-training the sentence encoder on a large unsupervised corpus yielded better accuracy than only pre-training word embeddings. Also, predicting the next token turned out to be a worse auxiliary objective than reconstructing the sentence itself, as the LSTM hidden state was only responsible for a rather short-term objective.

\subsection{Deep generative models}
Recent success in generating realistic images has driven a series of efforts on applying deep generative models to text data. The promise of such research is to discover rich structure in natural language while generating realistic sentences from a latent code space. In this section, we review recent research on achieving this goal with variational autoencoders (VAEs)~\cite{kingma2013auto} and generative adversarial networks (GANs)~\cite{goodfellow2014generative}.

\begin{figure}[h]
	\includegraphics[width=0.5\linewidth]{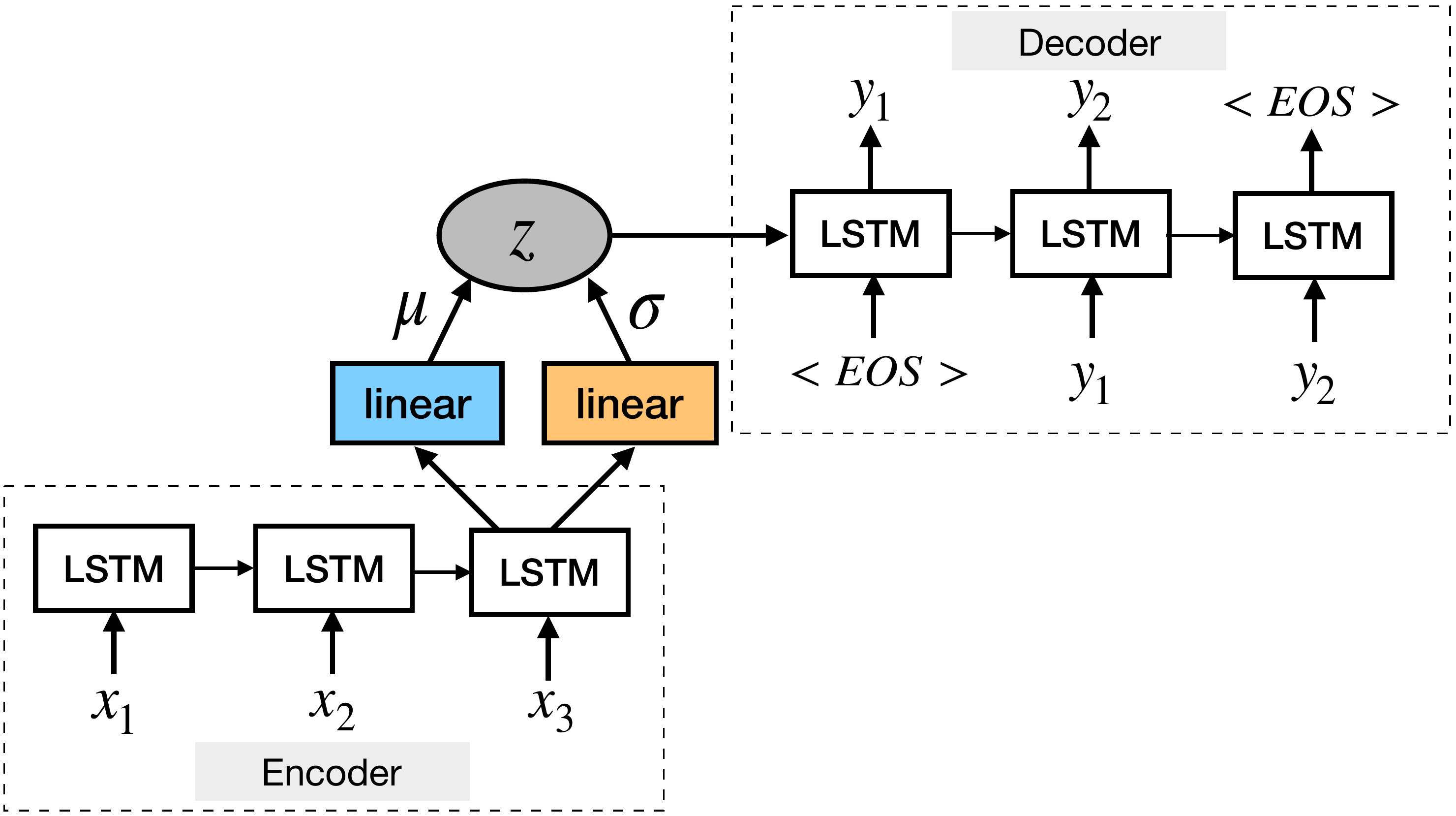}
	\centering
	\caption{RNN-based VAE for sentence generation (Figure source:~\citet{bowman2015generating})}\label{fig:VAE}
\end{figure}

Standard sentence autoencoders, as in the last section, do not impose any constraint on the latent space, as a result, they fail when generating realistic sentences from arbitrary latent representations~\cite{bowman2015generating}. The representations of these sentences may often occupy a small region in the hidden space and most of regions in the hidden space do not necessarily map to a realistic sentence~\cite{zhanggenerating}. They cannot be used to assign probabilities to sentences or to sample novel sentences~\cite{bowman2015generating}.

The VAE imposes a prior distribution on the hidden code space which makes it possible to draw proper samples from the model. It modifies the autoencoder architecture by replacing the deterministic encoder function with a learned posterior recognition model. The model consists of encoder and generator networks which encode data examples to latent representation and generate samples from the latent space, respectively. It is trained by maximizing a variational lower bound on the log-likelihood of observed data under the generative model. 

\citet{bowman2015generating} proposed an RNN-based variational autoencoder generative model that incorporated distributed latent representations of entire sentences (Fig.~\ref{fig:VAE}). Unlike vanilla RNN language models, this model worked from an explicit global sentence representation. Samples from the prior over these sentence representations produced diverse and well-formed sentences.

\citet{hu2017controllable} proposed generating sentences whose attributes are controlled by learning disentangled latent representations with designated semantics. The authors augmented the latent code in the VAE with a set of structured variables, each targeting a salient and independent semantic feature of sentences. The model incorporated VAE and attribute discriminators, in which the VAE component trained the generator to reconstruct real sentences for generating plausible text, while the discriminators forced the generator to produce attributes coherent with the structured code. When trained on a large number of unsupervised sentences and a small number of labeled sentences, \citet{hu2017controllable} showed that the model was able to generate plausible sentences conditioned on two major attributes of English: tense and sentiment.

GAN is another class of generative model composed of two competing networks. A generative neural network decodes latent representation to a data instance, while the discriminative network is simultaneously taught to discriminate between instances from the true data distribution and synthesized instances produced by the generator. GAN does not explicitly represent the true data distribution $p(x)$. 

\citet{zhanggenerating} proposed a framework for employing LSTM and CNN for adversarial training to generate realistic text. The latent code $z$ was fed to the LSTM generator at every time step. CNN acted as a binary sentence classifier which discriminated between real data and generated samples. One problem with applying GAN to text is that the gradients from the discriminator cannot properly back-propagate through discrete variables. In~\cite{zhanggenerating}, this problem was solved by making the word prediction at every time ``soft'' at the word embedding space. \citet{yu2017seqgan} proposed to bypass this problem by modeling the generator as a stochastic policy. The reward signal came from the GAN discriminator judged on a complete sequence, and was passed back to the intermediate state-action steps using Monte Carlo search.

The evaluation of deep generative models has been challenging. For text, it is possible to create oracle training data from a fixed set of grammars and then evaluate generative models based on whether (or how well) the generated samples agree with the predefined grammar~\cite{rajeswar2017adversarial}. Another strategy is to evaluate BLEU scores of samples on a large amount of unseen test data. The ability to generate similar sentences to unseen real data is considered a measurement of quality~\cite{yu2017seqgan}. 

\section{Memory-augmented Networks}\label{sec:7}
The attention mechanism stores a series of hidden vectors of the encoder, which the decoder is allowed to access during the generation of each token. Here, the hidden vectors of the encoder can be seen as entries of the model's ``internal memory''. Recently, there has been a surge of interest in coupling neural networks with a form of memory, which the model can interact with.

\begin{figure}[t]
 \includegraphics[scale=0.35]{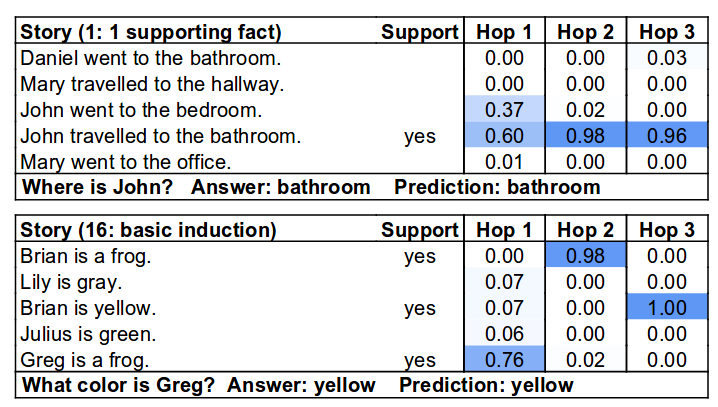}
 \centering
 \caption{Multiple supporting facts were retrieved from the memory in order to answer a specific question using an attention mechanism. The first hop uncovered the need for additional hops (Figure source:~\citet{sukhbaatar2015end})}\label{fig:hops}
\end{figure}

In~\cite{weston2014memory}, the authors proposed memory networks for QA tasks. In synthetic QA, a series of statements (memory entries) were provided to the model as potential supporting facts to the question. The model learned to retrieve one entry at a time from memory based on the question and previously retrieved memory. In large-scale realistic QA, a large set of commonsense knowledge in the form of (subject, relation, object) triples were used as memory.
\citet{sukhbaatar2015end} extended this work and proposed end-to-end memory networks, where memory entries were retrieved in a ``soft'' manner with attention mechanism, thus enabling end-to-end training. Multiple rounds (hops) of information retrieval from memory were shown to be essential to good performance and the model was able to retrieve and reason about several supporting facts to answer a specific question (Fig.~\ref{fig:hops}). \citet{sukhbaatar2015end} also showed a special use of the model for language modeling, where each word in the sentence was seen as a memory entry. With multiple hops, the model yielded results comparable to deep LSTM models.

Furthermore, dynamic memory networks (DMN)~\cite{kumar2015ask} improved upon previous memory-based models by employing neural network models for input representation, attention, and answer mechanisms. The resulting model was applicable to a wide range of NLP tasks (QA, POS tagging, and sentiment analysis), as every task could be cast to the $<$memory, question, answer$>$ triple format. \citet{xiong2016dynamic} applied the same model to visual QA and proved that the memory module was applicable to visual signals. 

\section{Performance of different models on different NLP tasks}\label{sec:8}
We summarize the performance of a series of deep learning methods on standard datasets developed in recent years on some major NLP topics. Our goal is to show the readers common datasets used in the community and state-of-the-art results with different models.

\subsection{POS tagging}
The WSJ-PTB (the Wall Street Journal part of the Penn Treebank Dataset) corpus contains 1.17 million tokens and has been widely used for developing and evaluating POS tagging systems. 
\citet{gimenez2004fast} employed one-against-all SVM based on manually-defined features within a seven-word window, in which some basic n-gram patterns were evaluated to form binary features such as: ``previous word is {\it the}'', "two preceding tags are DT NN", etc. One characteristic of the POS tagging problem was the strong dependency between adjacent tags. With a simple left-to-right tagging scheme, this method modeled dependencies between adjacent tags only by feature engineering. 
In an effort to reduce feature engineering, \citet{collobert2011natural} relied on only word embeddings within the word window by a multi-layer perceptron. Incorporating CRF was proven useful in~\cite{collobert2011natural}. 
\citet{santos2014learning} concatenated word embeddings with character embeddings to better exploit morphological clues. In~\cite{santos2014learning}, the authors did not consider CRF, but since word-level decision was made on a context window, it can be seen that dependencies were modeled implicitly. \citet{huang2015bidirectional} concatenated word embeddings and manually-designed word-level features and employed bidirectional LSTM to model arbitrarily long context. A series of ablative analysis suggested that bi-directionality and CRF both boosted performance. \citet{andor2016globally} showed a transition-based approach that produces competitive result with a simple feed-forward neural network.
When applied to sequence tagging tasks, DMNs~\cite{kumar2015ask} essentially allowed for attending over the context multiple times by treating each RNN hidden state as a memory entry, each time focusing on different parts of the context.

\begin{table*}[h]
	\centering
	\caption{\textbf{POS tagging}}
	\label{tab:POS}
	\begin{tabular}{|c|c|c|}
		\hline
		\textbf{Paper}   & \textbf{Model}      & \textbf{WSJ-PTB (per-token accuracy \%)} \\ \hline
		\citet{gimenez2004fast}  & SVM with manual feature pattern & 97.16    \\ \hline
		\citet{collobert2011natural}  & MLP with word embeddings + CRF   & 97.29    \\ \hline
		\citet{santos2014learning}   & MLP with character+word embeddings  & 97.32    \\ \hline
		\citet{huang2015bidirectional}  & LSTM & 97.29    \\ \hline
		\citet{huang2015bidirectional}  & Bidirectional LSTM & 97.40    \\ \hline
		\citet{huang2015bidirectional}  & LSTM-CRF & 97.54    \\ \hline
		\citet{huang2015bidirectional}  & Bidirectional LSTM-CRF & 97.55    \\ \hline
    \citet{andor2016globally}  & Transition-based neural network & 97.45    \\ \hline
		\citet{kumar2015ask} & DMN   & 97.56    \\ \hline
	\end{tabular}
\end{table*}

\subsection{Parsing}
There are two types of parsing: dependency parsing, which connects individual words with their relations, and constituency parsing, which iteratively breaks text into sub-phrases.
Transition-based methods are a popular choice since they are linear in the length of the sentence. The parser makes a series of decisions that read words sequentially from a buffer and combine them incrementally into the syntactic structure~\cite{chen2014fast}. 
At each time step, the decision is made based on a stack containing available tree nodes, a buffer containing unread words and the obtained set of dependency arcs. \citet{chen2014fast} modeled the decision making at each time step with a neural network with one hidden layer. The input layer contained embeddings of certain words, POS tags and arc labels, which came from the stack, the buffer and the set of arc labels. 

\citet{tu2015context} extended the work of~\citet{chen2014fast} by employing a deeper model with 2 hidden layers. However, both~\citet{tu2015context} and~\citet{chen2014fast} relied on manual feature selecting from the parser state, and they only took into account a limited number of latest tokens. \citet{dyer2015transition} proposed stack-LSTMs to model arbitrarily long history. The end pointer of the stack changed position as the stack of tree nodes could be pushed and popped. \citet{zhou2017neural} integrated beam search and contrastive learning for better optimization.

Transition-based models were applied to constituency parsing as well. \citet{zhu2013fast} based each transition action on features such as the POS tags and constituent labels of the top few words of the stack and the buffer. By uniquely representing the parsing tree with a linear sequence of labels, \citet{vinyals2015grammar} applied the seq2seq learning method to this problem.

\begin{table*}[h]
	\centering
	\caption{\textbf{Parsing} \small(UAS/LAS = Unlabeled/labeled Attachment Score; WSJ = The Wall Street Journal Section of Penn Treebank)}
	\label{tab:parsing}
	\begin{tabular}{|c|c|c|c|c|}
		\hline
		\textbf{Parsing type}    & \textbf{Paper}        & \textbf{Model}      & \textbf{WSJ} \\ \hline
		\multirow{3}{*}{Dependency Parsing} &~\citet{chen2014fast} & Fully-connected NN with features including POS & 91.8/89.6 (UAS/LAS) \\ \cline{2-4} 
		&~\citet{weiss2015structured} & Deep fully-connected NN with features including POS & 94.3/92.4 (UAS/LAS)      \\ \cline{2-4} 
		&~\citet{dyer2015transition}     & Stack-LSTM      & 93.1/90.9 (UAS/LAS)      \\ \cline{2-4} 
		&~\citet{zhou2017neural}     & Beam contrastive model      & 93.31/92.37 (UAS/LAS)\\ \hline

		\multirow{3}{*}{Constituency Parsing}         &~\citet{petrov2006learning} & Probabilistic context-free grammars (PCFG)  & 91.8 (F1 Score) 
		
		\\ \cline{2-4} 
		&~\citet{socher2011parsing}  & Recursive neural networks  & 90.29 (F1 Score)
		\\ \cline{2-4}
		&~\citet{zhu2013fast}  & Feature-based transition parsing  & 91.3 (F1 Score)
		\\ \cline{2-4} 
		&~\citet{vinyals2015grammar}     & seq2seq learning with LSTM+Attention    & 93.5 (F1 Score)                   \\ \hline
		
	\end{tabular}
\end{table*}

\begin{table*}[h]
	\centering
	\caption{\textbf{Named-Entity Recognition}}
	\label{tab:NER}
	\begin{tabular}{|c|c|c|}
		\hline
		\textbf{Paper}      &\textbf{ Model }    & \textbf{CoNLL 2003 (F1 \%)} \\ \hline

		\citet{collobert2011natural}    & MLP with word embeddings+gazetteer  & 89.59  \\ \hline
		\citet{passos2014lexicon}    & Lexicon Infused Phrase Embeddings & 90.90  \\ \hline
		\citet{chiu2015named}    & Bi-LSTM with word+char+lexicon embeddings  & 90.77  \\ \hline
		\citet{luo2015joint}    & Semi-CRF jointly trained with linking & 91.20  \\ \hline
		\citet{lample2016neural} & Bi-LSTM-CRF with word+char embeddings    & 90.94  \\ \hline
		\citet{lample2016neural} & Bi-LSTM with word+char embeddings    & 89.15  \\ \hline
		\citet{strubell2017fast} & Dilated CNN with CRF    & 90.54 \\ \hline
	\end{tabular}
\end{table*}

\subsection{Named-Entity Recognition}
CoNLL 2003 has been a standard English dataset for NER, which concentrates on four types of named entities: people, locations, organizations and miscellaneous entities. NER is one of the NLP problems where lexicons can be very useful. \citet{collobert2011natural} first achieved competitive results with neural structures augmented by gazetteer features. \citet{chiu2015named} concatenated lexicon features, character embeddings and word embeddings and fed them as input to a bidirectional LSTM. On the other hand, \citet{lample2016neural} only relied on character and word embeddings, with pre-training embeddings on large unsupervised corpora, they achieved competitive results without using any lexicon. Similar to POS tagging, CRF also boosted the performance of NER, as demonstrated by the comparison in~\cite{lample2016neural}. Overall, we see that bidirectional LSTM with CRF acts as a strong model for NLP problems related to structured prediction.

\citet{passos2014lexicon} proposed to modify skip-gram models to better learn entity-type related word embeddings that can leverage information from relevant lexicons. \citet{luo2015joint} jointly optimized the entities and the linking of entities to a KB. \citet{strubell2017fast} proposed to use dilated convolutions, defined over a wider effective input width by skipping over certain inputs at a time, for better parallelization and context modeling. The model showed significant speedup while retaining accuracy.  

\subsection{Semantic Role Labeling}
Semantic role labeling (SRL) aims to discover the predicate-argument structure of each predicate in a sentence. For each target verb (predicate), all constituents in the sentence which take a semantic role of the verb are recognized. Typical semantic arguments include Agent, Patient, Instrument, etc., and also adjuncts such as Locative, Temporal, Manner, Cause, etc.~\cite{zhou2015end}. Table~\ref{tab:SRL} shows the performance of different models on the CoNLL 2005 and 2012 datasets.

Traditional SRL systems consist of several stages: producing a parse tree, identifying which parse tree nodes represent the arguments of a given verb, and finally classifying these nodes to determine the corresponding SRL tags. Each classification process usually entails extracting numerous features and feeding them into statistical models~\cite{collobert2011natural}. 

Given a predicate, \citet{tackstrom2015efficient} scored a constituent span and its possible role to that predicate with a series of features based on the parse tree. They proposed a dynamic programming algorithm for efficient inference. \citet{collobert2011natural} achieved comparable results with a convolution neural networks augmented by parsing information provided in the form of additional look-up tables. \citet{zhou2015end} proposed to use bidirectional LSTM to model arbitrarily long context, which proved to be successful without any parsing tree information. \citet{he2017deep} further extended this work by introducing highway connections~\cite{srivastava2015training}, more advanced regularization and ensemble of multiple experts.

\begin{table*}[h]
	\centering
	\caption{\textbf{Semantic Role Labeling}}
	\label{tab:SRL}
	\begin{tabular}{|c|c|c|c|}
		\hline
		\textbf{Paper }                          & \textbf{Model}      & \textbf{CoNLL2005 (F1 \%)} & \textbf{CoNLL2012 (F1 \%)} \\ \hline
		
		\citet{collobert2011natural}                          & CNN with parsing features     & 76.06    &     \\ \hline
		\citet{tackstrom2015efficient}           & Manual features with DP for inference       & 78.6    & 79.4    \\ \hline
		\citet{zhou2015end}         & Bidirectional LSTM & 81.07    & 81.27    \\ \hline
		\citet{he2017deep} &  Bidirectional LSTM with highway connections      & 83.2    & 83.4    \\ \hline
	\end{tabular}
\end{table*}

\subsection{Sentiment Classification}
The Stanford Sentiment Treebank (SST) dataset contains sentences taken from the movie review website Rotten Tomatoes. It was proposed by~\citet{pang2005seeing} and subsequently extended by~\citet{socher2013recursive}. The annotation scheme has inspired a new dataset for sentiment analysis, called CMU-MOSI, where sentiment is studied in a multimodal setup~\cite{zadeh2016multimodal}. 

\cite{socher2013recursive} and~\cite{tai2015improved} were both recursive networks that relied on constituency parsing trees. Their difference shows the effectiveness of LSTM over vanilla RNN in modeling sentences. On the other hand, tree-LSTM performed better than linear bidirectional LSTM, implying that tree structures can potentially better capture the syntactical property of natural sentences. \citet{yu2017refining} proposed to refine pre-trained word embeddings with a sentiment lexicon, observing improved results based on \cite{tai2015improved}.

\citet{kim2014convolutional} and~\citet{KalchbrennerACL2014} both used convolutional layers. The model~\cite{kim2014convolutional} was similar to the one in Fig.~\ref{fig:CNN}, while~\citet{KalchbrennerACL2014} constructed the model in a hierarchical manner by interweaving k-max pooling layers with convolutional layers. 

\begin{table*}[h]
	\centering
	\caption{\textbf{Sentiment Classification} \small(SST-1 = Stanford Sentiment Treebank, fine-grained 5 classes~\citet{socher2013recursive}; SST-2: the binary version of SST-1; Numbers are accuracies (\%))
	}
	\label{tab:SA}
	\begin{tabular}{|c|c|c|c|}
		\hline
		\textbf{Paper }         & \textbf{Model }    & \textbf{SST-1} & \textbf{SST-2} \\ \hline
		\citet{socher2013recursive} & Recursive Neural Tensor Network & 45.7 & 85.4 \\ \hline
		\citet{kim2014convolutional}   & Multichannel CNN   & 47.4 & 88.1 \\ \hline
		\citet{KalchbrennerACL2014}    & DCNN with k-max pooling  & 48.5 & 86.8 \\ \hline
		\citet{tai2015improved}    & Bidirectional LSTM  & 48.5 & 87.2 \\ \hline
		\citet{le2014distributed}    & Paragraph Vector   & 48.7 & 87.8 \\ \hline
		\citet{tai2015improved}    & Constituency Tree-LSTM  & 51.0 & 88.0 \\ \hline
    \citet{yu2017refining}    & Tree-LSTM with refined word embeddings  & 54.0 & 90.3 \\ \hline
		\citet{kumar2015ask} & DMN  & 52.1 & 88.6 \\ \hline
		
	\end{tabular}
\end{table*}

\subsection{Machine Translation}
The phrase-based SMT framework~\cite{koehn2003statistical} factorized the translation model into the translation probabilities of matching phrases in the source and target sentences. \citet{cho2014learning} proposed to learn the translation probability of a source phrase to a corresponding target phrase with an RNN encoder-decoder. Such a scheme of scoring phrase pairs improved translation performance. \citet{sutskever2014sequence}, on the other hand, re-scored the top 1000 best candidate translations produced by an SMT system with a 4-layer LSTM seq2seq model. Dispensing the traditional SMT system entirely, \citet{wu2016google} trained a deep LSTM network with 8 encoder and 8 decoder layers with residual connections as well as attention connections. \citet{wu2016google} then refined the model by using reinforcement learning to directly optimize BLEU scores, but they found that the improvement in BLEU scores by this method did not reflect in human evaluation of translation quality. Recently, \citet{gehring2017convolutional} proposed a CNN-based seq2seq learning model for machine translation. The representation for each word in the input is computed by CNN in a parallelized style for the attention mechanism. The decoder state is also determined by CNN with words that are already produced. \citet{Vaswani2017attention} proposed a self-attention-based model and dispensed convolutions and recurrences entirely.

\begin{table*}[h]
	\centering
	\caption{\textbf{Machine translation} \small(Numbers are BLEU scores)}
	\label{tab:MT}
	\begin{tabular}{|c|c|c|c|}
		\hline
		\textbf{Paper}   & \textbf{Model }      & \textbf{WMT2014 English2German} & \textbf{WMT2014 English2French} \\ \hline
		\citet{cho2014learning}   & Phrase table with neural features    &     & 34.50     \\ \hline
		\citet{sutskever2014sequence} & Reranking phrase-based SMT best list with LSTM seq2seq &     & 36.5     \\ \hline
		\citet{wu2016google}   & Residual LSTM seq2seq + Reinforcement learning refining   & 26.30     & 41.16     \\ \hline
		\citet{gehring2017convolutional}   & seq2seq with CNN     & 26.36     & 41.29     \\ \hline
		\citet{Vaswani2017attention} & Attention mechanism     & 28.4     & 41.0     \\ \hline
	\end{tabular}
\end{table*}

\subsection{Question answering}
QA problems take many forms. Some rely on large KBs to answer open-domain questions, while others answer a question based on a few sentences or a paragraph (reading comprehension). For the former, we list (see Table~\ref{tab:QA}) several experiments conducted on a large-scale QA dataset introduced by~\cite{fader2013paraphrase}, where 14M commonsense knowledge triples are considered as the KB. Each question can be answered with a single-relation query. For the latter, we consider (see Table~\ref{tab:QA}) (1) the synthetic dataset of \emph{bAbI}~\cite{weston2015towards}, which requires the model to reason over multiple related facts to produce the right answer. It contains 20 synthetic tasks that test a model's ability to retrieve relevant facts and reason over them. Each task focuses on a different skill such as \emph{basic coreference} and \emph{size reasoning}. (2) The Stanford Question Answering Dataset (\emph{SQuAD})~\cite{rajpurkar2016squad}, consisting of 100,000+ questions posed by crowdworkers on a set of Wikipedia articles. The answer to each question is a segment of text from the corresponding article.

The central problem of learning to answer single-relation queries is to find the single supporting fact in the database. \citet{fader2013paraphrase} proposed to tackle this problem by learning a lexicon that maps natural language patterns to database concepts (entities, relations, question patterns) based on a question paraphrasing dataset. \citet{bordes2014open} embedded both questions and KB triples as dense vectors and scored them with inner product. 

\citet{weston2014memory} took a similar approach by treating the KB as long-term memory, while casting the problem in the framework of a memory network.
On the \emph{bAbI} dataset, \citet{sukhbaatar2015end} improved upon the original memory networks model~\cite{weston2014memory} by making the training procedure agnostic of the actual supporting fact, while~\citet{kumar2015ask} used neural sequence models (GRU) instead of neural bag-of-words models as in~\cite{sukhbaatar2015end} and~\cite{weston2014memory} to embed memories.

For models on the \emph{SQuAD} dataset, the goal is to determine the start point and end point of the answer segment. \citet{chen2017reading} encoded both the question and the words in context using LSTMs and used a bilinear matrix for calculating the similarity between the two. \citet{shen2017reasonet} proposed Reasonet, a model that read a document repeatedly with attention on different parts each time until a satisfying answer is found. \citet{yu2018qanet} replaced RNNs with convolution and self-attention for encoding the question and the context with significant speed improvement.

\begin{table*}[h]
	\centering
	\caption{\textbf{Question answering}}
	\label{tab:QA}
	\begin{tabular}{|c|c|c|c|c|}
		\hline
		\textbf{Paper}       & \textbf{Model}   & \textbf{bAbI (Mean accuracy \%)} & \textbf{Farbes (Accuracy \%)} & \textbf{SQuAD (EM/F1 \%)} \\ \hline
		\citet{fader2013paraphrase} & Paraphrase-driven lexicon learning &   & 0.54 &    \\ \hline
		\citet{bordes2014open}       & Weekly supervised embedding &   & 0.73  &   \\ \hline
		\citet{weston2014memory}       & Memory networks   & 93.3  & 0.83  &   \\ \hline
		\citet{sukhbaatar2015end}     & End-to-end memory networks  & 88.4  &   &   \\ \hline
		\citet{kumar2015ask}     & DMN & 93.6  &   &   \\ \hline
        \citet{chen2017reading}     & Document Reader  &   &   &  70.0/79.0 \\ \hline
        \citet{shen2017reasonet}     & ReasoNet  &   &   &  69.1/78.9  \\ \hline
		\citet{yu2018qanet}     & QAnet  &   &   &  76.2/84.6 \\ \hline
	\end{tabular}
\end{table*}

\subsection{Dialogue Systems}
Two types of dialogue systems have been developed: generation-based models and retrieval-based models. 

In Table~\ref{tab:DS}, the Twitter Conversation Triple Dataset is typically used for evaluating generation-based dialogue systems, containing 3-turn Twitter conversation instances. One commonly used evaluation metric is BLEU~\cite{papineni2002bleu}, although it is commonly acknowledged that most automatic evaluation metrics are not completely reliable for dialogue evaluation and additional human evaluation is often necessary. \citet{ritter2011data} employed the phrase-based statistical machine translation (SMT) framework to ``translate'' the message to its appropriate response. \citet{sordoni2015neural} reranked the 1000 best responses produced by SMT with a context-sensitive RNN encoder-decoder framework, observing substantial gains. \citet{li2015diversity} reported results on replacing the traditional maximum log likelihood training objective with the maximum mutual information training objective, in an effort to produce interesting and diverse responses, both of which are tested on a 4-layer LSTM encoder-decoder framework.

The response retrieval task is defined as selecting the best response from a repository of candidate responses. Such a model can be evaluated by the recall1@$k$ metric, where the ground-truth response is mixed with $k-1$ random responses. 
The Ubuntu dialogue dataset was constructed by scraping multi-turn Ubuntu trouble-shooting dialogues from an online chatroom~\cite{lowe2015ubuntu}. \citet{lowe2015ubuntu} used LSTMs to encode the message and response, and then inner product of the two sentence embeddings is used to rank candidates. 

\citet{zhou2016multi} proposed to better exploit the multi-turn nature of human conversation by employing the LSTM encoder on top of sentence-level CNN embeddings, similar to~\cite{serban2016building}. \citet{dodge2015evaluating} cast the problem in the framework of a memory network, where the past conversation was treated as memory and the latest utterance was considered as a ``question'' to be responded to. The authors showed that using simple neural bag-of-word embedding for sentences can yield competitive results.

\begin{table*}[h]
	\centering
	\caption{\textbf{Dialogue systems}}
	\label{tab:DS}
	\begin{tabular}{|c|c|c|c|}
		\hline
		\textbf{Paper}   & \textbf{Model}     & \begin{tabular}[c]{@{}c@{}}\textbf{Twitter Conversation} \\\textbf{ Triple Dataset (BLEU)}\end{tabular} & \begin{tabular}[c]{@{}c@{}}\textbf{Ubuntu Dialogue} \\ \textbf{Dataset (recall 1@10 \%)}\end{tabular} \\ \hline
		\citet{ritter2011data} & SMT     & 3.60           &           \\ \hline
		\citet{sordoni2015neural} & SMT+neural reranking   & 4.44           &           \\ \hline
		\citet{li2015diversity}   & LSTM seq2seq    & 4.51           &           \\ \hline
		\citet{li2015diversity}   & LSTM seq2seq with MMI objective  & 5.22           &           \\ \hline
		\citet{lowe2015ubuntu}  & Dual LSTM encoders for semantic matching &           & 55.22          \\ \hline
		\citet{dodge2015evaluating} & Memory networks    &           & 63.72          \\ \hline
		\citet{zhou2016multi}  & Sentence-level CNN-LSTM encoder    &           & 66.15          \\ \hline
	\end{tabular}
\end{table*}

\subsection{Contextual Embeddings} \label{sec:contextualembedingsresults}

In this section, we explore some of the recent results based on contextual embeddings as explained in section~\ref{sec:contextualembeddings}. ELMo has contributed significantly towards the recent advancement of NLP. In various NLP tasks, ELMo outperformed state of the art by significant margin (Table \ref{tab:elmo}). However, latest mode BERT surpass ELMo to establish itself as the state-of-the-art in multiple tasks as summarized in Table~\ref{tab:bert}.

\begin{table}[h]
\centering
\caption{Comparison of ELMo + Baseline with the previous state of the art (SOTA) on various NLP tasks. The table has been adapted from \citep{peters2018deep}. SOTA results have been taken from \citep{peters2018deep}; SQUAD~\citep{rajpurkar2016squad}: QA task; SNLI~\citep{bowman2015large}: Stanford Natural Language Inference task; SRL~\citep{zhou2015end}: Semantic Role Labelling; Coref~\citep{pradhan2012conll}: Coreference Resolution; NER~\citep{tjong2003introduction}: Named Entity Recognition; SST-5~\citep{socher2013recursive}: Stanford Sentiment Treebank 5-class classification;}
\begin{tabular}{|l|l|l|l|l|l|}
\hline
\textbf{Task} & \textbf{Previous SOTA}         & \textbf{\begin{tabular}[c]{@{}l@{}}Previous\\ SOTA Results\end{tabular}} & \textbf{\begin{tabular}[c]{@{}l@{}}Baseline\end{tabular}} & \textbf{\begin{tabular}[c]{@{}l@{}}ELMo + \\ Baseline\end{tabular}} & \textbf{\begin{tabular}[c]{@{}l@{}}Increase\\ (Absolute/Relative)\end{tabular}} \\ \hline
SQuAD         & \citet{liu}    & 84.4                                                                & 81.1                                                             & 85.8                                                                & 4.7 / 24.9\%                                                                    \\ \hline
SNLI          & \citet{chen}   & 88.6                                                                & 88.0                                                             & \SI{88.7 \pm 0.17}                                                         & 0.7 / 5.8\%                                                                     \\ \hline
SRL           & \citet{he}     & 81.7                                                                & 81.4                                                             & 84.6                                                                & 3.2 / 17.2\%                                                                    \\ \hline
Coref         & \citet{lee}    & 67.2                                                                & 67.2                                                             & 70.4                                                                & 3.2 / 9.8\%                                                                     \\ \hline
NER           & \citet{peters} & \SI{91.93 \pm 0.19}                                                        & 90.15                                                            & \SI{92.22 \pm 0.10}                                                        & 2.06 / 21\%                                                                     \\ \hline
SST-5         & \citet{mccann} & 53.7                                                                & 51.4                                                             & 54.7 ± 0.5                                                          & 3.3 / 6.8\%                                                                     \\ \hline
\end{tabular}
\label{tab:elmo}
\end{table}

\begin{table}[h]
\centering
\begin{tabular}{|l|l|l|}
\hline
\textbf{Task} & \begin{tabular}[c]{@{}l@{}}\textbf{BiLSTM+}\\ \textbf{ELMo+Attn}\end{tabular} & \textbf{BERT} \\ \hline
QNLI & 79.9 & 91.1 \\ \hline
SST-2 & 90.9 & 94.9 \\ \hline
STS-B & 73.3 & 86.5 \\ \hline
RTE & 56.8 & 70.1 \\ \hline
SQuAD & 85.8 & 91.1 \\ \hline
NER & 92.2 & 92.8 \\ \hline
\end{tabular}
\caption{QNLI~\citep{wang2018glue}: Question Natural Language Inference task; SST-2~\citep{socher2013recursive}: Stanford Sentiment Treebank binary classification; STS-B~\citep{cer2017semeval}:  Semantic Textual Similarity Benchmark; RTE~\citep{bentivogli2009fifth}: Recognizing Textual Entailment; SQUAD~\citep{rajpurkar2016squad}: QA task; NER~\citep{tjong2003introduction}: Named Entity Recognition.}
\label{tab:bert}
\end{table}

\section{Conclusion}\label{sec:9}
Deep learning offers a way to harness large amount of computation and data with little engineering by hand~\cite{lecun2015deep}. With distributed representation, various deep models have become the new state-of-the-art methods for NLP problems. Supervised learning is the most popular practice in recent deep learning research for NLP. In many real-world scenarios, however, we have unlabeled data which require advanced unsupervised or semi-supervised approaches. In cases where there is lack of labeled data for some particular classes or the appearance of a new class while testing the model, strategies like zero-shot learning should be employed. These learning schemes are still in their developing phase but we expect deep learning based NLP research to be driven in the direction of making better use of unlabeled data. We expect such trend to continue with more and better model designs. We expect to see more NLP applications that employ reinforcement learning methods, e.g., dialogue systems. 
We also expect to see more research on multimodal learning~\cite{baltruvsaitis2017multimodal} as, in the real world, language is often grounded on (or correlated with) other signals.

Finally, we expect to see more deep learning models whose internal memory (bottom-up knowledge learned from the data) is enriched with an external memory (top-down knowledge inherited from a KB). Coupling symbolic and sub-symbolic AI will be key for stepping forward in the path from NLP to natural language understanding. Relying on machine learning, in fact, is good to make a `good guess' based on past experience, because sub-symbolic methods encode correlation and their decision-making process is probabilistic. Natural language understanding, however, requires much more than that. To use Noam Chomsky's words, ``you do not get discoveries in the sciences by taking huge amounts of data, throwing them into a computer and doing statistical analysis of them: that's not the way you understand things, you have to have theoretical insights''.
\appendices

\bibliographystyle{IEEEtranN}
\bibliography{bibexport}

\end{document}